\documentclass[lettersize,journal]{IEEEtran}

\usepackage[margin=0.7in]{geometry}

\usepackage{dblfloatfix}

\usepackage{colortbl}
\usepackage[x11names,table]{xcolor}
\usepackage{tikz}
\usetikzlibrary{automata, positioning, matrix, fit, arrows} 
\usetikzlibrary{backgrounds}
\usetikzlibrary{snakes}
\usepackage{mytikz-uml}
\tikzumlset{font=\sffamily}
\usepackage{sansmathfonts} 
\usepackage{url}

\usepackage{ulem}

\usepackage{bold-extra}

\usepackage{pgf}

\usepackage{2mil-masPaper}
\usepackage{fancyvrb}

\usepackage{graphicx}
\usepackage{xspace}

\usepackage{soul}

\usepackage{longtable}
\usepackage{tabularx}
\usepackage{booktabs}

\pgfdeclareimage[width=8cm]{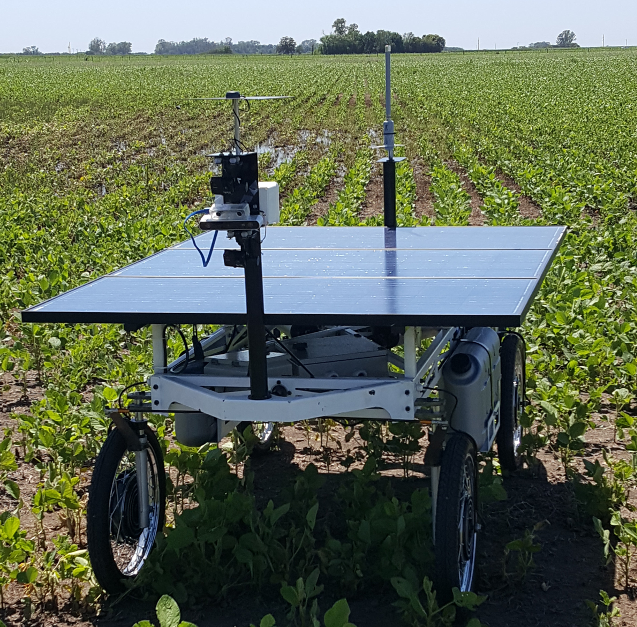}{./robot}
\pgfdeclareimage[width=7cm]{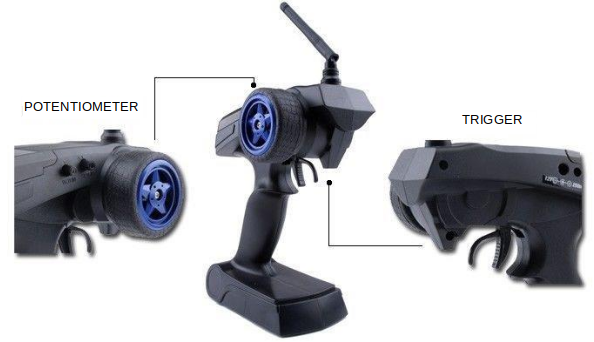}{./CR}
\pgfdeclareimage[width=8cm]{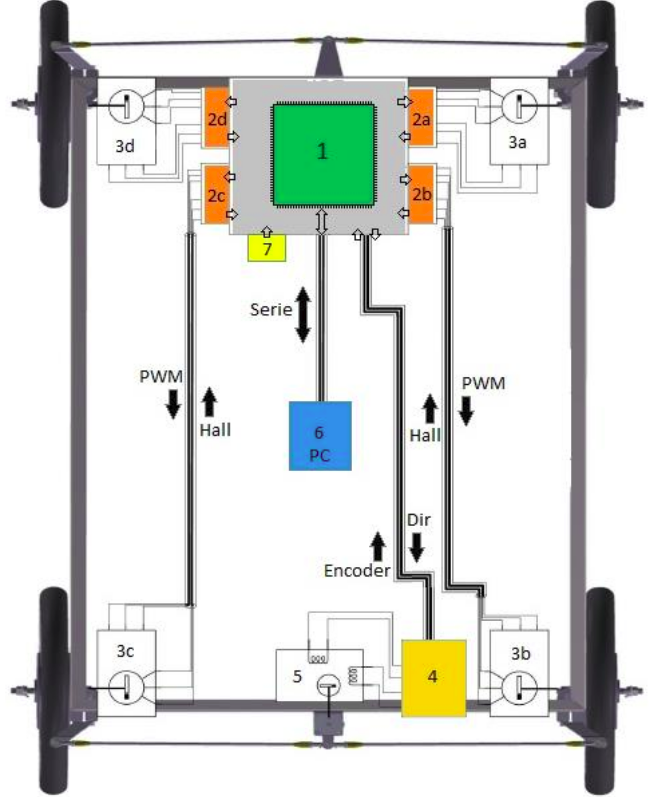}{./robotHard}
\pgfdeclareimage[width=2\linewidth]{documentation}{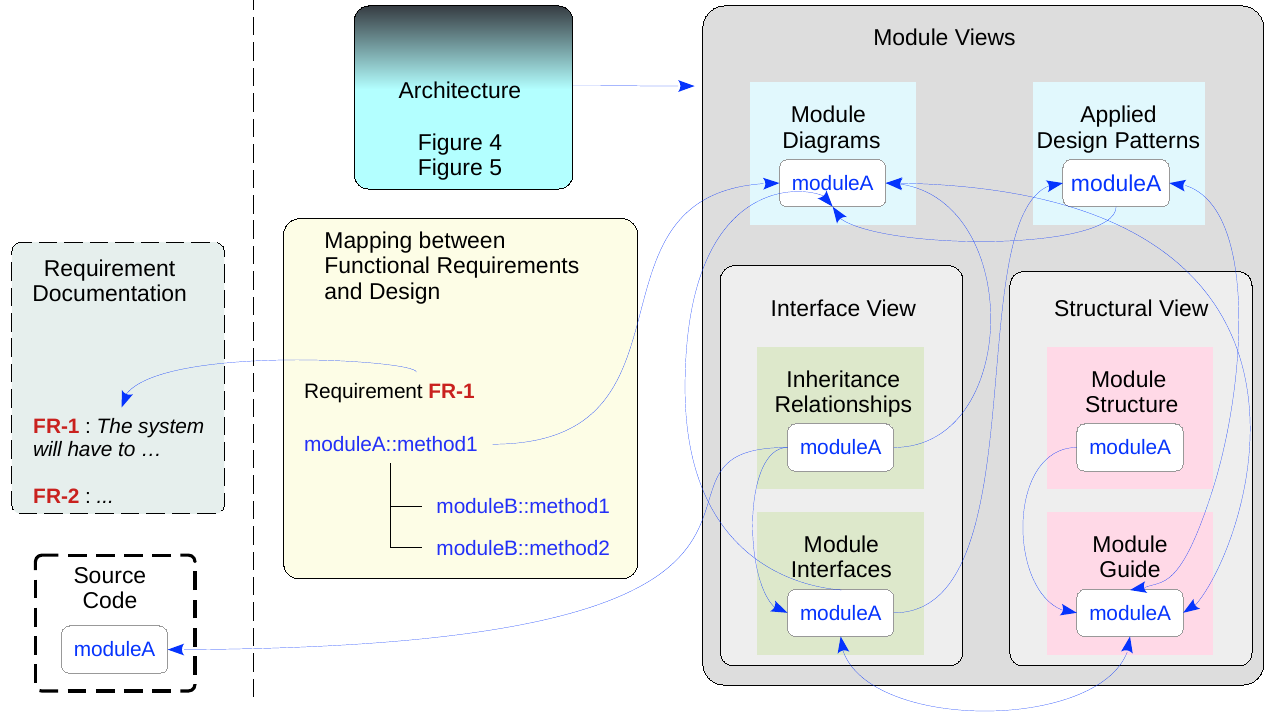}

\newcommand{\sfNoSep}[1]{\mbox{\textsf{#1}}}
\newcommand{\mtNoSep}[1]{\sfNoSep{#1}}

\newcommand{\mdNoSep}[1]{\textbf{\sfNoSep{#1}}}
\newcommand{\abstNoSep}[1]{\textit{\sfNoSep{#1}}}
\newcommand{\mdAbstMt}[2]{\mbox{\textsf{\textit{#1}::#2}}}

\newcommand{\mdMt}[2]{\mbox{\textsf{\textbf{#1}::#2}}}

\newcommand{\ms}{ms\xspace}
\newcommand{\rpm}{rpm\xspace}
\newcommand{\scNoSep}[1]{\mbox{\textsc{#1}}}
\newcommand{\cmpNoSep}[1]{{\sf\scNoSep{#1}}}

\newcommand{\robot}{WR\xspace}

\title{Reusability and Modifiability in Robotics Software  {\LARGE (Extended Version)}}
\author{%
\IEEEauthorblockN{Laura Pomponio} \\
\IEEEauthorblockA{{\small CIFASIS (CONICET-UNR), Argentina -- Email: \texttt{pomponio@cifasis-conicet.gov.ar}}}
\\[2mm]
\IEEEauthorblockN{Maximiliano Cristiá} \\
\IEEEauthorblockA{{\small Universidad Nacional de Rosario (UNR) and CIFASIS (CONICET-UNR), Argentina -- Email: \texttt{cristia@cifasis-conicet.gov.ar}}}
\\[2mm]
\IEEEauthorblockN{Estanislao Ruiz Sorazábal} \\
\IEEEauthorblockA{{\small Universidad Nacional de Rosario (UNR), Argentina -- Email: \texttt{estanislao.ruiz@gmail.com}}}
\\[2mm]
\IEEEauthorblockN{Maximiliano García} \\
\IEEEauthorblockA{{\small CIFASIS (CONICET-UNR), Argentina -- Email: \texttt{mgarcia@cifasis-conicet.gov.ar}}}
}

\begin{document}
\maketitle

\begin{abstract}
We show the design of the software of the microcontroller unit of a weeding robot based on the Process Control architectural style and design patterns. The design consists of 133 modules resulting from using 8 design patterns for a total of 30 problems. As a result the design yields more reusable components and an easily modifiable and
extensible program. Design documentation is also presented. Finally, the implementation (12 KLOC of C++ code) is empirically evaluated to prove that the design does not produce an inefficient implementation.
\end{abstract}

\section{Architectural Styles and Design Patterns}

Software architecture and design are considered essential artefacts to achieve important software quality attributes such as modifiability, reusability, maintainability, etc. \cite{ShawGarlan1996,ghezzi2003,bass2003,DBLP:books/daglib/0030743}. Robot software is frequently implemented in a few functions, where each one consists of a lot of lines of code and a vast number of nested conditional sentences \cite{dornaRobot, erdos}.
In some cases, the code is separated in smaller modules but carrying out a functional decomposition. The functional decomposition criterion for software design has been shown to produce poor results when the mentioned attributes are important \cite{Parnas1972}. Indeed, software built according to this design criterion is difficult to maintain and its components are less reusable. This paper shows 
how established trends in software architecture and design, such as \textit{architectural styles} and \textit{design patterns} \cite{Gamma:1995:DPE:186897,DBLP:books/lib/BuschmannHS07,ShawGarlan1996}, can be systematically applied to robot software to produce reusable components, and easily modifiable and extensible programs. 
For this purpose, the MicroController Unit (MCU) software architecture and design of a weeding robot are presented.

\subsection*{Design for change in robot  software}

In order to achieve software modifiability, reusability and maintainability, software designers must think in terms of \textit{likely changes} or \textit{design for change} \cite{Parnas1972,ShawGarlan1996,ghezzi2003,bass2003,DBLP:books/daglib/0030743}. 
In the case of a robot, its behavioural objectives, its hardware or its control algorithms, among others, will probably change over time. Consequently, modifications in the code will be required. Thus, if the robot software was not designed anticipating likely changes, these modifications can entail a great effort at the moment of carrying out the required changes.

A \textit{software product line} (or \textit{family of programs}) is a collection of similar software systems built, in part, from a shared set of software assets \cite{Parnas02,DBLP:books/daglib/0019719}. This set is the set of reusable components; the components outside it usually differ in implementation details (e.g. different algorithms or  interaction with different sensors) which is a particular kind of likely changes. 

The development of a family of robots can easily be thought as a software product line.
For instance, all the robots in the family perform similar (but not equal) functions over different hardware platforms. Thus, likely changes are the fundamentals to produce such a family.
Therefore, proper software architecture and design sensibly contribute to software modifiability, reusability and maintainability, and to the development of software product lines. However, these quality attributes might compromise performance. 
For example, in a layered architecture, the higher layers perform high-level functions while the lower layers have access to the hardware. Then, a high-level function need to go through intermediate layers before reaching the hardware. Consequently, this could introduce performance penalties, but several parts of the software would be more reusable. Depending on the particular features of each system, engineers can select architectural styles and design patterns that best adapt to each circumstance. 

This work has two main goals. One is to show that design patterns and architectural styles are key tools to build robot software easily implementing required changes. The other goal is to show empirical evidence that using such tools does not necessarily compromise system performance. Besides, the work aims at fostering some important design documents.

\section{Software Design in Robotics}

We were unable to find works focusing or describing the software design of robotics software.
In general, articles focus on the results obtained from the software or how a new algorithm or technique is implemented but not on how the software was designed. This is replied in embedded software, as well.
Usually, when there exists some reference to the defined software design, it is presented as a flowchart describing different functional steps, suggesting a lack of a proper modularity or a functional division of code. Some examples are the MCU software designs shown in \cite{Zhang2009, ChangSheng2012, YiNan2016}. 

Some works try to improve this approach by proposing designs in which some Software Engineering (SE) principles are applied.
The design of an autonomous agricultural mobile-robot, called Agrobot, is presented in 
\cite{Dumus2015}.
The authors show the Agrobot software structure where the mainboard software is designed in three levels. 
Although each level is described only in functional terms, this article is one of the few that would seem to consider some modularity and abstraction in the software design.

According to \cite{Kortenkamp2016B}, hierarchical and modular decomposition is a common, generally desirable feature of robot architectures. However, the authors agree that there exists a debate about the criteria in order to put into practice such decomposition. According to them, architectural styles can be used in order to address each particular situation. 
In \cite{MaryShaw94,ShawGarlan1996}, some architectural styles for robotics systems are discussed. Particularly, through the classic example of a car's cruise velocity, the authors show how an object-oriented approach is not sufficient to make it clear key aspects of the system; while an architectural style based on a control loop revels essential characteristics leading to better design decisions.

Recently, the Robotics community has begun to show interest in SE in order to improve robot software development.
In \cite{Brugali2009} the authors wonder about the role that software plays in robot development and they point out that there is no systematic approach in order to design reusable robotics software components and to implement flexible and modular code. As a result, developers must constantly rewrite code. 
The authors overview some Robotics works which have applied some SE techniques and principles, and they manifest that adopting such practices has arisen as a necessity in robot development.

In this context, framework-based development has appeared as an alternative to facilitate and to improve developers' tasks \cite{BrugaliEtAl2007,Wienke2012}. Several robotic-specific frameworks and tools have been developed in order to give response to some typical difficulties found in robotic software \cite{Kortenkamp2016}. Although these frameworks are useful instruments with which a developer can count on, they are not sufficient to deal with difficulties that arise when software has to be modified.

An interesting project which applies SE techniques is Boss \cite{Baker2009}. 
Boss was the winner of The Urban Challenge (2007), an autonomous vehicle competition sponsored by the U.S. Defence Advanced Research Project Agency (DARPA). The Boss control software was designed by using design patterns in an incremental development process, in which requirements were not completely defined a priori. Although software architecture and design appears as topics in the article, its goal is not to focus on a methodology  to build resusable and modifiable robotic software. With the intention of contributing in this sense, our article presents a methodological approach in order to design software, that anticipates changes, based on architectural styles and design patterns.

\section{Design for Change}

Maintainability and reusability are some desirable quality attributes of a software artefact. Maintainability refers to the evolving and corrective capacity of software; that is, the possibility of implementing new requirements, modifying some functionalities or correcting defects with a reasonable effort. 
Reusability aims at the possibility of, maybe with little changes, using a software component as part of another system.

Changes could occur in the software platform on which our programs run, in data structures representing information, in hardware components, in algorithms and so on. Change is a distinguishing characteristic of SE \cite{Parnas1972,ghezzi2003,bass2003}. Thus, software should be designed in a way that each change can be introduced at a low cost and without compromising the software structural integrity. Implementing likely changes should be more advantageous than developing the software again from scratch. 

Accomplishing quality attributes such as these ones, requires software development to be based on some SE best-practices: \textit{anticipation of change}, \textit{modularity}, \textit{information hiding} and the \textit{open-closed design principle}  \cite{ghezzi2003,Parnas1972,DBLP:books/ph/Meyer97}.

Anticipation of change is a general principle, to which the other concepts are strongly related. According to Ghezzi et al., during the design activity engineers anticipate likely changes that the software may undergo during its lifetime and, as a result, they produce a software design that will accommodate these changes easily \cite{ghezzi2003}. 
The techniques used to achieve this objective are denominated as \textit{design for change}.

The modularity principle applied to software design means to conceive a complex system as a set of simpler modules which cooperate among them to fulfill the system's functionality. Each module has certain properties and responsibilities in the system. In turn, modules relate to each other in a clearly defined way. 
One of the benefits of modularity is the possibility of modifying a system by modifying only a small number of its modules. The difficult part of applying modularity is to find out the modularisation criterion that maximizes reusability, maintainability and modifiability.

Precisely, anticipation of change together with \textit{information hiding} have been shown to be a modularisation criterion producing software featuring those quality attributes \cite{Parnas1972,ghezzi2003,bass2003}. Information hiding states that each anticipated, likely change must be implemented in a single module  that implements only that. The implementation of a likely change is \textit{hidden} behind the module's interface. For example, think in a sensor returning the sensed value as a signed integer. A likely change is to replace that sensor by another one that could return the sensed value as two unsigned integers: one storing the value and the other its sign. If the system uses the value in several places, changing the sensor could entail the modification of several functions. Conversely, if the interaction with the sensor is concealed to a single module presenting an abstract interface to the rest of the system, changing the sensor entails the modification of a single module.

Adding new functionality by adding new modules is even better than adding it by modifying existing modules. Indeed, every time an existing module is modified chances are of introducing errors. Modifying an error-free module is not the best idea, as modified modules should be thoroughly verified. The open-closed design principle states that systems should be open to extensions but closed to modifications \cite{DBLP:books/ph/Meyer97}. In this way new functionality is added by extending the system with new modules. For example, consider the sensor mentioned above. Assume that after some experiments with the robot it is determined that the sensor produces a noisy signal. Implementing a filter to reduce noise can be done as part of the module hiding the interaction with the sensor (call it \textbf{A}). However, a better choice is to \textit{wrap} \textbf{A} with another module (call it \textbf{B}) having the same interface than \textbf{A}. \textbf{B} implements just the noise filter, and delegates in \textbf{A} the interaction with the sensor. That is, \textbf{A} and \textbf{B} each hides exactly one likely change. Given that both modules have the same interface, the rest of the system will not know that now is interacting with \textbf{B}. Furthermore, if the sensor is replaced by a new, less noisy one, \textbf{B} can be simply removed from the system.

Architectural styles and design patterns are techniques based on the mentioned SE best-practices allowing to decompose a system into components and modules, and to achieve a design for change \cite{ShawGarlan1996,Gamma:1995:DPE:186897,DBLP:books/lib/BuschmannHS07}. Each architectural style and design pattern has been conceived as to accommodate a set of likely changes in a way that all the nice quality attributes are preserved.
Both architectural styles and design patterns provide a variety of decomposition perspectives which adjust to different characteristics and which can be selected for software decomposition and organization.
Some architectural styles were proposed to be used in a handful of application domains and, symmetrically, some application domains have de-facto reference architectures which are known to preserve quality attributes \cite{ShawGarlan1996,DBLP:books/lib/BuschmannHS07}. When this is the case there is no need to consider likely changes because styles will cover recurring likely changes coming from the dynamics of the application domain.
Design patterns enjoy similar properties in a smaller, more detailed scale. If some functionality and design restrictions fit in the description of a design pattern, then some likely changes will be covered by the pattern itself. Each design pattern was defined to deal with certain recurrent likely changes occurring in program development.

Architectural styles are applied first as they provide a coarse structural decomposition (high-level design or (sub)system design). Design patterns are applied in a second phase as they provide a detailed division (detailed design or module design).

The greatest advantage of using styles and patterns is, perhaps, that they have been proved over the years in many application domains and systems, and by many software engineers across the globe. Styles and patterns do deliver modifiability, reusability and maintainability.

In summary, the design methodology goes as follows:
\begin{enumerate}
\item Consider likely changes for the robot's software.
\item Choose the architectural styles or reference architectures proposed for robotics software or related domains (e.g. process control, artificial intelligence, etc.). If there are no such styles or architectures, choose the architectural styles that best accommodate those likely changes. 
\item Select design patterns for fine-tuning the design. In this stage, likely changes drive design pattern selection.
\end{enumerate}

\section{A Weeding Robot}
\begin{figure}
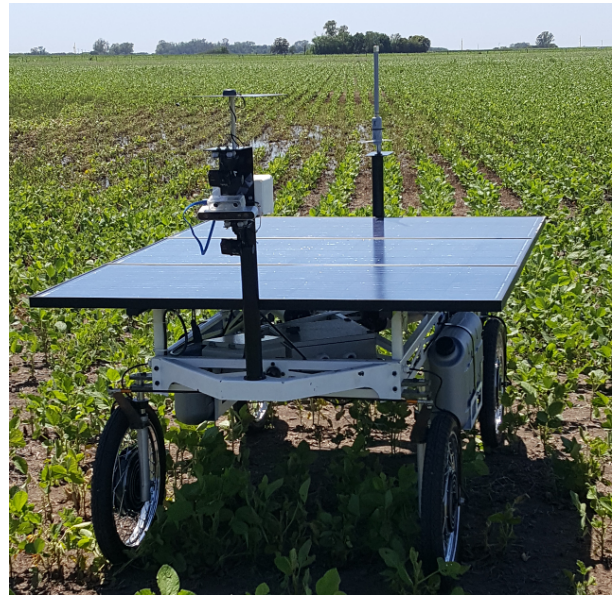

\centering
\pgfuseimage{robot}
\caption{The weeding Robot (\robot)}
\label{fig:AR}\hypertarget{fig:AR}{}
\end{figure}

We applied the proposed methodology to the MCU software design of a weeding robot (WR) (Figure \ref{fig:AR}). This robot is a prototype conceived for precision agriculture whose goal is weed automatic removal. By means of artificial vision and image processing, the \robot is able to move along furrows and to distinguish weeds from crops. Once weed is detected, this robot is capable of locally applying the suitable agrochemical. 

The code and documentation of the project is available in a public repository \cite{WRobotMCU}.

\subsection{\robot Hardware}

\begin{figure}
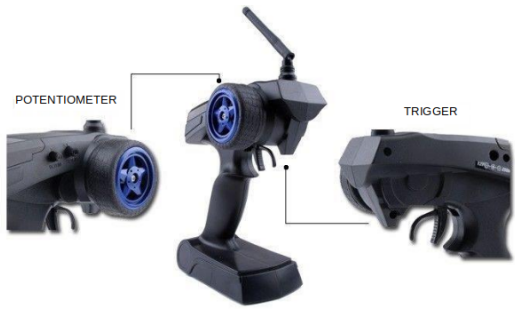

\centering
\pgfuseimage{CR}
\caption{Remote control}
\label{fig:CR}
\end{figure}

The robot counts on four wheels and a steering device to make them turn.
Each wheel has Hall sensors, which allow to measure its position and velocity, and an associated current measurement system. 
The steering device allows to determine its angular position at each moment. 
Wheels and steering device can be remotely operated by means of a remote control (RC) (Figure \ref{fig:CR}) capable of sending direction and velocity signals  to a radio-frequency (RF) receiver module, situated on the robot. 
Besides, a computer (PC) situated on the robot sends orders to the steering device and to the wheels for \robot autonomous navigation.
Orders coming from both the RC and the PC are processed by a MCU located on the robot.

\begin{figure}
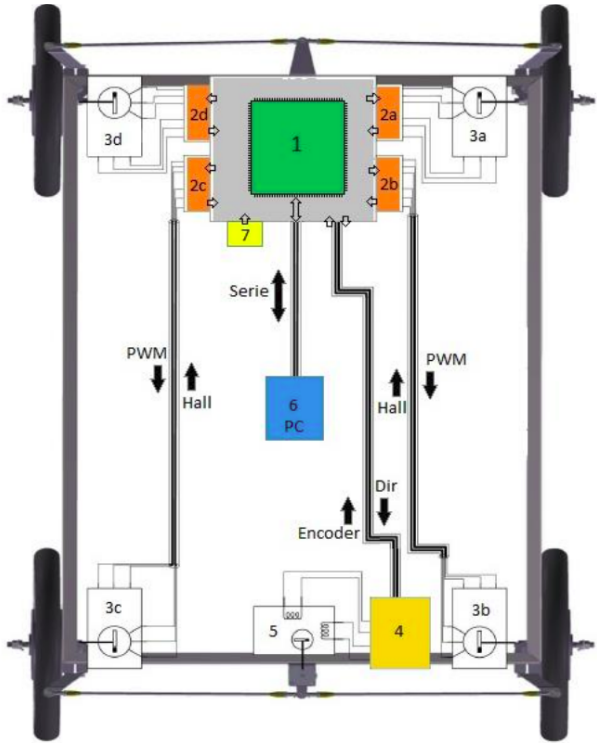

\centering
\pgfuseimage{robotHard}
\caption{Hardware of the weeding robot}
\label{fig:robotHard}
\end{figure}

Figure \ref{fig:robotHard} depicts the \robot hardware principal components, on which the software described here operates. Components are identified with numbers and their descriptions are as follows.
1) Main cabinet responsible for controlling the behaviour of wheels and steering device. This component acts as main controller and contains the MCU that executes the software. 
2a--3d) 
Brushless DC motors and drivers associated with the wheels. 
4--5) Stepper motor and driver of the robot steering device. 
6) PC from which the MCU will receive orders through serial communication. 
7) RF module by means of which the MCU will receive orders from the RC.

\subsection{\label{functionality}\robot MCU functionality}

The software running on the MCU continuously carries out execution  cycles in order to control the behaviour of the steering device and the wheels.

During each execution cycle, orders coming from the PC or the RC are processed by the MCU. These orders could specify the desired angular position for the steering device, as well as the desired velocity, tension or current for the wheels.
In addition, the MCU will have to make readings and measurements of actual values of angular position, current and velocity.

From desired values (set-points or reference values) and measured values, certain calculations are completed in order to determine turn direction and tension to be applied to the wheels. The purpose is to keep measured values close to desired values. Once this is accomplished, the corresponding signals will be sent to the steering device and the wheels. Afterwards, the MCU sends measured values and calculations to the PC. Then, a new control cycle takes place.

It is noteworthy that a process control model underlies the described functionality. The goal of this controller is to maintain the values of process measurable properties equal or sufficiently near to the received reference values.

Finally, the \robot MCU software implements all the basic, low-level functions an operating system would normally provide (e.g., serial communication).

\section{MCU Software Architecture}

Robotics software is typically structured in a three-tier (3T) architecture \cite{Kortenkamp2016B}, depicted in Figure \ref{fig:arqCtrlWD}. Then, we use this as the reference architecture for the \robot. In turn, the \robot MCU software is clearly placed in the \textsc{Behavioural control} tier due to the functionality it implements (cf. Section \ref{functionality}) and the function assigned to that tier by the 3T architecture.

However, if the functionality of the \robot MCU software is implemented according to a simple, functional design, likely changes will be hard to implement. In other words, the program implementing that functionality is as large and complex as to deserve a thoughtful design. Therefore, we need to choose an architectural style to design the MCU software. As explained in Section \ref{functionality}, the MCU software should implement a \textit{process control} system for the wheels and the steering device. Shaw and Garlan \cite[Sections 2.8 and 3.4]{MaryShaw94} have proposed a style specifically tailored for process control systems, known simply as the Process Control architectural style. Clearly, unless unlikely conditions occur, this should be the style of choice, without even carrying out a thoroughly study on anticipated changes. The style has been defined considering likely changes identified in the process control domain. The authors argue that, given a system in the process control domain, an architecture based on Process Control is superior than those that can be obtained by applying standard object-oriented design criteria \cite[Sections 3.4.1 and 3.4.3]{MaryShaw94}.

\begin{figure}
\centering
\vspace{0.5cm}
\scalebox{0.7}{
\begin{tikzpicture}
\sf
\tikzstyle{moduloL}=[minimum width=1cm, minimum height=1.5cm,inner sep=1mm,above right,draw,align=center, font=\scshape] 

\tikzstyle{supest}=[rounded corners=1.5mm, minimum width=2cm,inner sep=3mm,draw,text width=2cm]

\tikzstyle{nombre}=[inner sep=0mm, font=\bfseries]

\tikzstyle{pipe}=[-latex,thick,line width=4pt]

\tikzstyle{nombreLogico}=[inner sep=0mm, font=\scshape, minimum width=1.5cm]

\tikzstyle{ctrl}=[shape=circle,draw,minimum width=1cm, inner sep=0, align=center,font=\sf\scshape];

\tikzstyle{sensor}=[draw=black,fill=red,circle, minimum width=0.2cm,after node path={(\tikzlastnode)  circle (0.5cm)}]

\node[ctrl, text width=1.9cm, outer color = orange,inner color=white] (0,0) (controlM){Main Control};
\node[ctrl, text width=1.9cm, below left=0.85cm and -0.36cm of controlM, ball color=yellow] (controlW3){};
\node[ctrl, text width=1.9cm, below left=0.9cm and -0.28cm of controlM, ball color=yellow] (controlW2){};
\node[ctrl, text width=1.9cm, below left=0.95cm and -0.19cm of controlM, ball color=yellow] (controlW1){};
\node[ctrl, text width=1.9cm, below left=1cm and -0.1cm of controlM, outer color=yellow!80!gray, inner color =white] (controlW){Wheel Control};
\node[ctrl, text width=1.9cm, below right=1cm and -0.1cm of controlM,outer color=yellow!70!black, inner color=white] (controlD){Direction Control};

\node[circle,draw=gray!70!white,line width=4pt,inner sep=-1mm, align=center, fit=(controlM)(controlW)(controlD)] (controlG){};

\node[moduloL, below right=2.15cm and -1.45cm of controlW, minimum width=2cm, fill = blue!60](wheel3){};
\node[moduloL, below right=2.2cm and -1.55cm of controlW, minimum width=2cm, fill = blue!60](wheel2){};
\node[moduloL, below right=2.25cm and -1.65cm of controlW, minimum width=2cm, fill = blue!60](wheel1){};
\node[moduloL, outer color=blue!60,inner color=white, below=2cm of controlW, minimum width=2cm](wheel){Wheel};

\node[moduloL, below=2cm of controlD, text width=1.7cm, outer color = violet!60!white,inner color=white](dir){Steering Device};
\node[rectangle, draw=gray!70!white,line width=4pt, inner sep=3mm, fit=(wheel)(dir)]{};
%
\draw[ball color=white,draw=red] node[sensor, below =1.2cm of wheel.-121]{};
\draw[ball color=white,draw=red] node[sensor, below =1.3cm of wheel.-124]{};
\draw[ball color=white,draw=red] node[sensor, below =1.4cm of wheel.-127]{};
\draw[ball color=white,draw=red] node[sensor, below =1.5cm of wheel.-130](sensorVel){};
\draw[ball color=white,draw=red] node[sensor, below=1.2cm of wheel.-41]{};
\draw[ball color=white,draw=red] node[sensor, below=1.3cm of wheel.-44]{};
\draw[ball color=white,draw=red] node[sensor, below=1.4cm of wheel.-47]{};
\draw[ball color=white,draw=red] node[sensor, below=1.5cm of wheel.-50](sensorCte){};
\draw[ball color=white,draw=red] node[sensor, below=1.5cm of dir](sensorDir){};

%
\node[below=3.7cm of wheel.-130](pto1){};
\node[below left=-0.2cm and -0.3cm of pto1]{velocity};
\node[below=3.7cm of wheel.-50](pto2){current};
\node[below=3.7cm of dir](pto3){position};
\draw[dashed, -latex](wheel.-130)--(pto1);
\draw[dashed, -latex](wheel.-50)--(pto2);
\draw[dashed, -latex](dir)--(pto3);
%

\node(outSensorVel)[left=0.05cm of sensorVel.west]{};
\draw[pipe] (outSensorVel) -| (-3.3,-4) |- (controlW.south west);
\node(outSensorCnt)[below=0.05cm of sensorCte.south]{};
\draw[pipe] (outSensorCnt) |- (-3.7,-9.7) |- (controlW.west);
\node(outSensorDir)[right=0.4cm of sensorDir.west]{};
\draw[pipe] (outSensorDir) -| (3.5,-4) |- (controlD.east);

\draw[-latex](controlW) edge (wheel);
\draw[-latex](controlD) edge (dir);

\draw[-latex](controlM) edge (controlW);
\draw[-latex](controlM) edge (controlD);
\node[below left=0.7cm and -1cm of controlW, text width=1.3cm]{tension};
\node[below right=0.7cm and -0.7cm of controlD, text width=1.3cm]{position};
\node[above =2cm of controlG.160](pv){velocity};
\node[above =1.5cm of controlG.120, text width=1.3cm](pt){tension};
\node[above =1.5cm of controlG.60](pc){current};
\node[above =2cm of controlG.20](pp){position};
\node[above left =1cm and 0.5cm of controlG.180, text width=1.5cm](ptime){clock pulse};

\draw[-latex](pv) edge (controlM.160);
\draw[-latex](pt) edge (controlM.120);
\draw[-latex](pc) edge (controlM.60);
\draw[-latex](pp) edge (controlM.20);
\draw[*-latex](ptime) edge (controlM.180);

\node[above left=4cm and 0cm of controlM,color= white](behavioural){\large\textsc{Behavioural control}};
\node[above left=6cm and 0cm of controlM, color=white](executive){\large\textsc{Executive}};
\node[above left=8cm and 0cm of controlM, color=white](planning){\large\textsc{Planning}};

\node[above =0.2cm of planning](pArriba){};
\node[below left=0.5cm and 3cm of planning](pIzq1){};
\node[below right=0.5cm and 6cm of planning](pDer1){};
\draw[loosely dashed, white, ultra thick] (pIzq1) -- (pDer1);

\node[below left=2.5cm and 3cm of planning](pIzq2){};
\node[below right=2.5cm and 6cm of planning](pDer2){};
\draw[loosely dashed, white, ultra thick] (pIzq2) -- (pDer2);

\begin{pgfonlayer}{background}

\node[top color=black!80!white, bottom color=cyan!5!white, middle color=cyan!50!white, rectangle, rounded corners,fit=(pArriba)(pIzq1)(pDer1)(pto3)]{};
\end{pgfonlayer}

\node[below left=13cm and 3cm of controlM](f11){};
\node[below left=13cm and 1cm of controlM](f12){};
\draw[*-latex] (f11) edge node[above](f1){event} (f12);

\node[below left=13cm and 0.2cm of controlM](f21){};
\node[below right=13cm and 0.2cm of controlM](f22){};
\draw[-latex] (f21) edge node[above, text width =2.5cm](f2){procedure call} (f22);

\node[below right=13cm and 1cm of controlM](f31){};
\node[below right=13cm and 3cm of controlM](f32){};
\draw[dashed,-latex] (f31) edge node[above, text width =1.5cm](f3){physical reponse}(f32);

\node[below =1.3cm of f11](f41){};
\node[below =1.3cm of f12](f42){};
\draw[pipe] (f41) edge node[above, text width =1.5cm](f4){pipe}(f42);

\node[shape=circle,draw=gray!60!white,line width=4pt,minimum width=1.2cm,below=1.5cm of f2,label={above,text width=1.7cm:control\\ algorithm}](c){};

\node[outer color=gray!50!white, inner color= white,shape=circle,draw,minimum width=1.2cm,below=1.5cm of f3,label={above, align=center,text width=2.5cm:control\\ sub-algorithm}](sc){};

\node[shape=rectangle,draw=gray!60!white,line width=4pt,minimum width=1.5cm,minimum height=0.8cm ,below=2.3cm of f4,label={above:process}](p){};

\node[outer color=gray!50!white, inner color= white, shape=rectangle,draw,minimum width=1.5cm,minimum height=0.8cm ,below=1cm of c,label={above, align=center:sub-process}](sp){};

\draw[ball color=white,draw=red] node[sensor, below=1.4cm of sc,label={[label distance=0.35cm] above,align=center:sensor}](s){};

\node[supest, fit=(f11)(f2) (f3) (f32)(f42)(c)(p)(sc)(sp)(s)]{};
\end{tikzpicture}
}
\caption{Process Control architecture of the robot MCU software}
\label{fig:arqCtrlWD}
\end{figure}

Software design based on the Process Control style proposes to clearly identify and distinguish: the \textit{process} to be controlled; the \textit{controlled variables}; the \textit{manipulated variables}; the \textit{set-points}; and the \textit{control algorithm}. All these elements can be found in the functional description given in Section \ref{functionality}, as shown below.

In the \robot, the behaviour of the steering device and wheels is considered as the \textit{process} to be controlled by the MCU software. However, this process is too complex as to be implemented as a single unit. 
Then, the \textit{process} is decomposed into a sub-process corresponding to the behaviour of the steering device, and a sub-processes corresponding to the behaviour of each wheel.
As prescribed by the Process Control style, these processes are modelled as software components (\cmpNoSep{Steering Device} and \cmpNoSep{Wheel}, Figure \ref{fig:arqCtrlWD})  capable of actuating on the corresponding hardware  in order to modify the controlled physical process.

\textit{Controlled variables} are those whose value the system must control; in this case, steering device \textit{position}, and wheel \textit{velocity} and \textit{current}. These variables are associated with \textit{sensors} (or measurement systems) which are modelled as software components that write sensed information to a software connector known as \textit{pipe}. A pipe connects two modules, a \textit{source} and a \textit{sink}. The source sends information to the sink but either of them knows where the information comes from or goes to. The source can be removed from the system or replaced by another module without modifications on the sink, and vice versa \cite[Section 2.2]{MaryShaw94,DBLP:books/lib/BuschmannHS07}. Then, control components will receive information and sensors will send it, unknowing who or to whom it was communicated.

\textit{Manipulated variables} are those whose values can be  modified by the system controller; that is, steering device position and wheel \textit{tension}. As Figure \ref{fig:arqCtrlWD} shows, these values can be altered by the controller. 

\textit{Set-points} are desired values for the controlled variables that the controller receives as input. In particular, these refer to position, velocity, current and tension.

The system \textit{controller} is a software component implementing the \textit{control algorithm} used to control the process under supervision. 
The controller receives set-points, reads the controlled variable values (velocity, current, position) from the pipes, and determines the values for the manipulated variables (tension, position). In this way, the controlled variables should be close to the set-points. 
Again, this controller is too complex to be implemented in a single unit. Then, in the \robot it is composed of a \cmpNoSep{Direction Control} responsible for controlling the \cmpNoSep{Steering Device}; the \cmpNoSep{Wheel Control} responsible for controlling a  \cmpNoSep{Wheel}; and a \cmpNoSep{Main Control}. The latter is responsible for receiving set-points and synchronising all actions that must be carried out by the other controllers in each control cycle. In particular, each control cycle will begin when a clock pulse occurs. Four instances of \cmpNoSep{Wheel Control} are executed concurrently.

The Process Control style allows to separate the process from the control algorithm. Therefore, the mechanism by means of which the physical process can be altered, manipulating some of its variables, is isolated from the control algorithm that establishes how such variables must be manipulated. 
In a similar way, the mechanism through which the values of controlled variables are obtained, is independent of the rest of the system. These architectural properties allow to accommodate likely changes inherent to the process control domain. For example, the control algorithm for the steering device can be improved independently of the other components;  changes in the sensors hardware do not impact beyond the components implementing the interaction with them; and so forth and so on.

\section{Detailed Design}\label{sec:detailedDsgn}

Once the architecture of the system has been laid down it needs to be refined into a detailed design. This refinement is carried out by decomposing the architectural elements into modules. Design patterns simplify the refinement in many cases.

\subsection{From architecture to detailed design}

The \textit{components} described in the MCU software architecture, as well as the \textit{connectors} between them (Figure \ref{fig:arqCtrlWD}), should be decomposed into modules.

In an object-oriented context, a module is usually implemented as a class. However, conceptually, a module is an Abstract Data Type (ADT) whose behaviour is defined through methods or functions declared in its interface, which is visible to other modules. Given that a module defines an ADT, a \textit{module instance} is an element of the defined ADT.
In object-oriented programming, a module instance would be an object of the corresponding class. Each module is characterised by a secret hidden to the other modules \cite{Parnas1972}. In turn, this secret corresponds to the implementation of a likely change. In this way, when this likely change needs to be realised, only the implementation of the module is modified. If the module's interface was correctly defined, the modification will not affect other modules.  

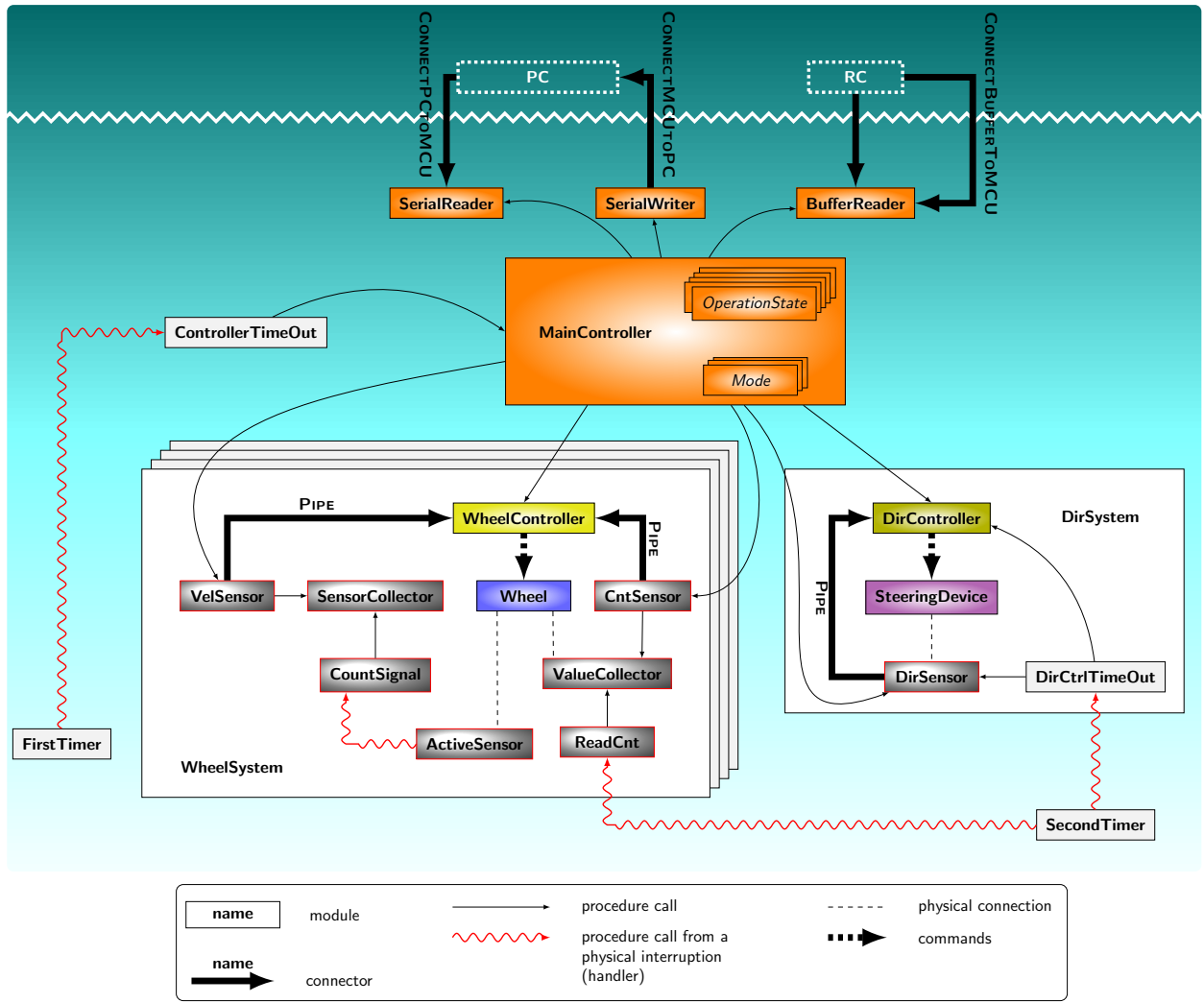
\begin{figure*}
\centering
\scalebox{0.65}{
\sf
\begin{tikzpicture}
\tikzstyle{modulo}=[minimum width=1cm,inner sep=2mm,above right,draw,align=center,minimum width=2cm, font=\bfseries]

\tikzstyle{supest}=[rounded corners=1.5mm, minimum width=2cm,inner sep=2mm,draw,text width=2cm]

\tikzstyle{nombre}=[inner sep=0mm, font=\bfseries]

\tikzstyle{pipe}=[-latex,thick,line width=4pt]

\tikzstyle{cmd}=[-latex,dashed,thick,line width=4pt]

\tikzstyle{modExt}=[minimum width=1cm,inner sep=2mm,above right,draw, dotted,line width=2pt,align=center,minimum width=2cm,color=gray, font=\bfseries]

\tikzstyle{flechaFisica}=[-latex,snake=coil,segment aspect=0, red, thick];


\draw[fill=gray!10](-6.1,-14.4) rectangle (6,-7.4);
\draw[fill=gray!10](-6.3,-14.6) rectangle (5.8,-7.6);
\draw[fill=gray!10](-6.5,-14.8) rectangle (5.6,-7.8);
\draw[fill=white](-6.7,-15) rectangle (5.4,-8);

\draw[fill=white](7,-13.2) rectangle (15.5,-8);


\draw[snake=zigzag, color=white, line width=2pt](-9.6,-0.5)--(16.3,-0.5);

\node[modExt, text width=3cm, color=white] (PC){PC};
\node[modExt, right= 4cm of PC, color=white](CR){RC};

\node[modulo, below left=2cm and -1cm of PC,outer color = orange,inner color=white](SerialReader){SerialReader};
\node[modulo, below right=2cm and -0.5cm of PC,outer color = orange,inner color=white](SerialWriter){SerialWriter};
\node[modulo, below=2cm of CR, outer color = orange,inner color=white](BufferReader) {BufferReader};

\draw[pipe] (PC.west) -| (SerialReader);
\node[nombre, above left=0.2cm and -0.9cm of SerialReader](ConnectPCtoMCU){\rotatebox{-90}{\textsc{ConnectPCtoMCU}}};

\draw[pipe] (SerialWriter) |- (PC.east);
\node[nombre, above right=0.2cm and -0.9cm of SerialWriter]{\rotatebox{-90}{\textsc{ConnectMCUtoPC}}};

\draw[pipe](CR)--(BufferReader);
\draw[pipe](CR.east)-| (11,-0.5) |-(BufferReader.east);
\node[nombre, above right=-0.6cm and 1.5cm of BufferReader]{\rotatebox{-90}{\textsc{ConnectBufferToMCU}}};

\node[nombre, below left=2.3cm and -1.2cm of SerialWriter](MCnombre){\hspace{0.5cm}MainController};
\node[modulo,outer color = orange,inner color=white,above right=0.6cm and 1cm of MCnombre, text width=2.5cm](O1){O};

\node[modulo,outer color = orange,inner color=white,below left=-0.55cm and -2.8cm of O1, text width=2.5cm]{O};
\node[modulo,outer color = orange,inner color=white,below left=-0.45cm and -2.7cm of O1, text width=2.5cm]{O};
\node[modulo,outer color = orange,inner color=white,below left=-0.35cm and -2.6cm of O1, text width=2.5cm]{O};
\node[modulo,outer color = orange,inner color=white,below left=-0.25cm and -2.5cm of O1, font=\itshape](OperationState){OperationState};


\node[modulo,outer color = orange,inner color=white,below right=0.4cm and 1.3cm of MCnombre](CRM){CR};
\node[modulo,outer color = orange,inner color=white,below left=-0.6cm and -1.9cm of CRM](PCM){PC};
\node[modulo,outer color = orange,inner color=white,below left=-0.5cm and -1.8cm of CRM, font=\itshape](Mode){Mode};

\node[modulo, below left = 3.5cm and -1.7cm of MCnombre, outer color=yellow!80!gray, inner color =white](WheelController){WheelController};

\node[modulo, below =1cm of WheelController, outer color=blue!60,inner color=white](Wheel){Wheel};

\node[modulo, right=0.5cm of Wheel, ball color=white,draw=red](CntSensor){CntSensor};

\node[modulo, below right=1cm and -0.6cm of Wheel, ball color=white,draw=red](ValueCollector){ValueCollector};

\node[modulo, below left=2.5cm and -1.2cm of Wheel, ball color=white,draw=red](ActiveSensor){ActiveSensor};

\node[modulo, below left=1cm and 3.8cm of WheelController, ball color=white,draw=red](VelSensor){VelSensor};

\node[modulo, right=0.7cm of VelSensor, ball color=white,draw=red](SensorCollector){SensorCollector};

\node[modulo, below=1cm of SensorCollector, ball color=white,draw=red](CountSignal){CountSignal};

\node[modulo, below=0.8cm of ValueCollector, ball color=white,draw=red](ReadCnt){ReadCnt};

\draw[cmd](WheelController) edge (Wheel);
\draw[dashed](Wheel.-150) edge (ActiveSensor.34);
\draw[dashed](Wheel.-28) edge (ValueCollector.164);
\draw[-latex](CntSensor) edge (ValueCollector.24);
\draw[-latex](ReadCnt) edge (ValueCollector);

\draw[-latex](VelSensor) edge (SensorCollector);
\draw[-latex](CountSignal) edge (SensorCollector);

\draw[flechaFisica](ActiveSensor.west) -| (CountSignal.-150);

\draw[pipe](VelSensor) |-(WheelController.west);
\node[nombre, above right=1.5cm and 0.5cm of VelSensor](pipe1){\textsc{Pipe}};

\draw[pipe](CntSensor.north) |- (WheelController.east);
\node[nombre, above right=0.5cm and -0.9cm of CntSensor](pipe2){\rotatebox{-90}{\textsc{Pipe}}};

\node[nombre,below right =3.2cm and -2cm of VelSensor](WheelSystem){WheelSystem};

\node[modulo, below right=3.5cm and 4.7cm of MCnombre, outer color=yellow!70!black, inner color=white] (DirController){DirController};

\node[modulo, below =1cm of DirController, outer color=violet!60,inner color=white] (SteeringDevice){SteeringDevice};

\node[modulo, below =1cm of SteeringDevice,ball color=white,draw=red] (DirSensor){DirSensor};

\node[modulo, right =1cm of DirSensor, fill=gray!10] (DirCtrlTimeOut){DirCtrlTimeOut};

\draw[cmd] (DirController) edge (SteeringDevice);
\draw[dashed] (SteeringDevice) -- (DirSensor);
\draw[-latex] (DirCtrlTimeOut) -- (DirSensor);
\draw[-latex] (DirCtrlTimeOut) edge [bend right] (DirController.-7);

\draw[pipe](DirSensor.west) -| (8,-10) |- (DirController.west);
\node[nombre,left=0.8 of SteeringDevice]{\rotatebox{-90}{\textsc{Pipe}}};

\node[nombre, right =1.5cm of DirController](DirSystem){DirSystem};

\node[modulo, below=2.5cm of DirCtrlTimeOut,fill=gray!10] (Timer){SecondTimer};
\draw[flechaFisica] (Timer) -- (DirCtrlTimeOut);
\draw[flechaFisica](Timer.west) -| (ReadCnt.south);


\node[modulo, left=6.5cm of ActiveSensor, fill=gray!10](Timer100){FirstTimer};

\node[modulo, left=4cm of MCnombre, fill=gray!10](ControllerTimeOut){ControllerTimeOut};

\draw[flechaFisica](Timer100) |- (ControllerTimeOut);


\node[right=0.4cm of DirCtrlTimeOut](p1fondo){};
\node[below=0.3cm of Timer](p2fondo){};

\begin{pgfonlayer}{background}
\node[top color=black!60!cyan, bottom color=cyan!5!white, middle color=cyan!50!white, rectangle, rounded corners,fit=(ConnectPCtoMCU)(Timer100) (p1fondo)(p2fondo)]{};

\node[draw,outer color = orange,inner color=white, rectangle, inner sep=2mm,fit=(O1) (Mode) (MCnombre)](MC){};
\end{pgfonlayer}

\draw[-latex](MC) edge[bend left] (BufferReader);
\draw[-latex](MC) edge (SerialWriter);

\draw[-latex](MC) edge[bend right] (SerialReader);

\draw[-latex](ControllerTimeOut) edge[out=90,bend left] (MC.west);

\draw[-latex](MC) edge[in=90,out=-140,looseness=0] (WheelController);
\draw[-latex](MC) edge[in=120,out=190,looseness=1.3]  (VelSensor);
\draw[-latex](MC) edge[in=0,out=-53,looseness=1]  (CntSensor);
\draw[-latex](MC) edge[in=90,out=-30,looseness=0]  (DirController);
\draw[-latex](MC) edge[in=200,out=-47,looseness=1.4]  (DirSensor);


\node[modulo, below left =12cm and 5cm of MCnombre](mod){name};
\node[right =0.5cm of mod](modDesc){module};

\node[below=1cm of mod.south west](x1){};
\node[below=1cm of mod.south east](x2){};
\draw[pipe](x1)--(x2);
\node[nombre, below =0.7cm of mod](pp){name};
\node[right=0.3cm of x2]{connector};


\node[below left =12cm and -1cm of MCnombre](x2a){};
\node[below right =12cm and -4.5cm of MCnombre](x1a){};

\draw[-latex](x1a)--(x2a);
\node[right=0.3cm of x2a]{procedure call};
\node[below=0.4cm of x1a](x1b){};
\node[below=0.4cm of x2a](x2b){};
\draw[flechaFisica](x1b)--(x2b);
\node[below right=-0.4cm and 0.3cm of x2b, text width=4cm](hand){procedure call from a physical interruption (handler)};
\node[below right =12cm and 3.5cm of MCnombre](x1c){};
\node[below right =12cm and 5cm of MCnombre](x2c){};
\draw[dashed](x1c)--(x2c);
\node[right=0.3cm of x2c](rf){physical connection};
\node[below=0.4cm of x1c](x1d){};
\node[below=0.4cm of x2c](x2d){};
\draw[cmd](x1d)--(x2d);
\node[right=0.3cm of x2d](comd){commands};

\node[supest, fit=(mod)(hand)(rf)(comd)]{};

\end{tikzpicture}
}
\caption{High level diagram of the MCU software}
\label{fig:diagGral}
\end{figure*}

Figure \ref{fig:diagGral} illustrates a refined view of the system where each architectural component has been refined into one or more collaborating modules. For example, an architectural component such as a \sfNoSep{velocity sensor} (Figure \ref{fig:arqCtrlWD}) becomes the modules \mdNoSep{VelSensor}, \mdNoSep{SensorCollector}, \mdNoSep{CountSignal} and \mdNoSep{ActiveSensor} (Figure \ref{fig:diagGral}). 
Each of the superposed boxes named \mdNoSep{WheelSystem}  represents a control thread in charge of one wheel. In other words, an instance of \mdNoSep{WheelSystem} is created when the system is initalised, is bound to the actual wheel and is ran in its own thread.
The modules are defined in terms of likely changes. For example, \mdNoSep{ActiveSensor} abstracts away the hardware of a Hall sensor. In this way, if the existing sensor is changed by a different model only the implementation of \mdNoSep{ActiveSensor} has to be changed; all the other modules will not be affected as they use the module's interface.
Table \ref{tab:compMod} details the relation between the two figures, where each high-level component is refined into a set of modules.

\begin{table}
\caption{Architectural components refined into modules}
\label{tab:compMod}
\centering
\begin{minipage}{\columnwidth}
\begin{tabularx}{\columnwidth}{l >{\raggedright\arraybackslash}X}
\toprule 
\multicolumn{1}{c}{\textsc{Component (Figure \ref{fig:arqCtrlWD})}} & \multicolumn{1}{c}{\textsc{Modules\footnote{Components might be refined into more modules than those shown here.} (Figure \ref{fig:diagGral})}} \\\midrule
&\\[-0.2cm]
\sfNoSep{velocity sensor}&\mdNoSep{VelSensor}, \mdNoSep{SensorCollector}, \mdNoSep{CountSignal}, \mdNoSep{ActiveSensor} \\
&\\[-0.2cm]
\sfNoSep{current sensor}&\mdNoSep{CntSensor}, \mdNoSep{ValueCollector}, \mdNoSep{ReadCnt} \\
&\\[-0.2cm]
\sfNoSep{position sensor}&\mdNoSep{DirSensor} \\
&\\[-0.2cm]
\cmpNoSep{Wheel}&\mdNoSep{Wheel} \\
&\\[-0.2cm]
\cmpNoSep{Steering Device}&\mdNoSep{SteeringDevice} \\
&\\[-0.2cm]
\cmpNoSep{Wheel Control}&\mdNoSep{WheelController} \\
&\\[-0.2cm]
\cmpNoSep{Direction Control}&\mdNoSep{DirController} \\
&\\[-0.2cm]
\cmpNoSep{Main Control}&\mdNoSep{MainController}, \abstNoSep{OperationState}, \abstNoSep{Mode}, \mdNoSep{SerialReader}, \mdNoSep{SerialWriter}, \mdNoSep{BufferReader} \\\bottomrule
\end{tabularx}
\end{minipage}
\end{table}

\subsection{Applying design patterns}

Below, the functionality of \cmpNoSep{Main Control} is described. After that, two different design approaches will be analysed.
The first one is based on a functional decomposition while the second one is based on design patterns. We will see how the latter accommodates likely changes more easily. In turn, we will see how \cmpNoSep{Main Control} is refined into several cooperative, smaller modules.

\subsubsection{Functionality of \cmpNoSep{Main Control}}

During each control cycle, \cmpNoSep{Main Control} reads a message (e.g., set-points or stops), synchronises the components that control the process and sends the results to the PC.

\begin{figure}
\centering
\scalebox{0.8}{
\begin{tikzpicture}
\small
\tikzstyle{state}=[shape=ellipse,minimum width=1cm, inner sep=4, align=center,font=\sffamily\bfseries];

\tikzstyle{stateWait}=[state, fill=yellow!70!orange!20!white,draw=yellow!70!orange];

\tikzstyle{stateWork}=[state, fill=green!70!orange!20!white,draw=green!70!orange];

\tikzstyle{stateReconnect}=[state,inner sep=3, fill=magenta!70!orange!10!white,draw=magenta!70!orange!];

\tikzstyle{statePC}=[state,inner sep=4, fill=SkyBlue1!80!white!30!,draw=SkyBlue1];

\tikzstyle{stateRC}=[state,inner sep=4, fill=violet!20!white!70!,draw=violet];

\tikzstyle{supest}=[rounded corners=5mm, minimum width=2cm,inner sep=2mm,draw]

\node[stateWait](w1){Waiting 1};

\node[stateWait, below = 0.7cm of w1](w2){Waiting 2};

\node[below = 0.7cm of w2](p1){};
\node[below = 1cm of p1](p2){};

\node[stateWait, below = 0.7cm of p2](wN){Waiting N};

\draw[-latex] (w1) edge (w2);
\node[below left=0.2cm and -1cm of w1,font=\sffamily]{noMessage};

\draw[-latex] (w2) edge (p1);
\node[below left=0.2cm and -1cm of w2,font=\sffamily]{noMessage};

\draw[dashed] (p1) edge (p2);
\draw[-latex] (p2) edge (wN);
\node[above left=0.2cm and -1cm of wN,font=\sffamily]{noMessage};


\node[stateWork, right = 2.7cm of w2](work){Working};

\node[above right =3cm and -1.5cm of work, font=\sffamily\bfseries](opSt){Operation States};

\node[stateReconnect, below = 2cm of work](reconnect){Reconnecting};

\draw[-latex](work) edge[in=45,out=90]  (w1);
\node[above left = 2.3cm and 0.9cm of work, font=\sffamily]{noMessage};

\draw[-latex](w1) edge[in=120,out=0,looseness=1]  (work);
\node[above left = 0.8cm and 2cm of work, text width=0.5cm, font=\sffamily]{newMessage/ read};

\draw[-latex](w2) edge[in=140,out=10, looseness=1]  (work);
\node[above left = -0.3cm and 2cm of work, text width=0.5cm, font=\sffamily]{newMessage/ read};

\draw[-latex](wN) edge[in=180,out=20, looseness=1]  (work);
\node[below left = 1cm and 1.7cm of work, text width=1cm, font=\sffamily]{newMessage/ read};

\draw[-latex](wN) edge[in=200,out=0, looseness=1]  (reconnect);
\node[below left = 0.6cm and 0.2cm of reconnect, font=\sffamily]{noMessage};

\draw[-latex](reconnect) edge  (work);
\node[above right = 0.6cm and -1.1cm of reconnect,text width=1cm, font=\sffamily]{newMessage/ read};

\draw[-latex](work) edge[out=60, in=20, looseness=4]  (work);
\node[above right = 0.5cm and -0.4cm of work, text width=1cm, font=\sffamily](pDer){newMessage/ read};

\node[right = 0.7cm of work](iniWork){};
\draw[*-latex](iniWork) edge  (work);

\draw[-latex](reconnect) edge[in=-40,out=250, looseness=4]  (reconnect);
\node[below right = 0.5cm and -0.9cm of reconnect,font=\sffamily]{noMessage};

\node[below left =0.7cm and 1cm of wN](e1){};
\node[below right =0.7cm and 7.2cm of wN](e2){};
\draw[dashed] (e1) edge (e2);

\node[below = 8.5cm of opSt, font=\sffamily\bfseries](rModes){Reading Modes};

\node[statePC,below left =1.3cm and 4cm of rModes](PC){PC};
\node[stateRC,below left =1.3cm and -1.5cm of rModes](RC){RC};

\draw[-latex](PC) edge[in=160,out=20](RC);
\node[above right =0.3cm and 0.1cm of PC, font=\sffamily]{noMessage $\land$ (in \textbf{Reconnecting})};

\draw[-latex](RC) edge[in=-20,out=200](PC);
\node[below right =0.3cm and 0.1cm of PC, font=\sffamily](pAbajo){noMessage $\land$ (in \textbf{Reconnecting})};

\node[right = 0.7cm of RC](iniRC){};
\draw[*-latex](iniRC) edge  (RC);

\draw[-latex](RC) edge[in=-20,out=-70, looseness=5](RC);
\node[below right= 0.3cm and -0.1cm of RC, font=\sffamily]{newMessage};

\draw[-latex](PC) edge[in=-150,out=-100, looseness=5.5](PC);
\node[below left= 0.4cm and -0.5cm of PC, font=\sffamily](pIzq){newMessage};

\node[right =0.2cm of e1](e11){};
\node[left= 0.2cm of e2](e22){};

\node[supest, fit=(pAbajo)(e11)(e22)(opSt)]{};

\end{tikzpicture}}
\caption{FSM describing internal states of \cmpNoSep{Main Control}. The dotted line means that the FSM execute concurrently.}
\label{fig:opStateFSM}
\end{figure}
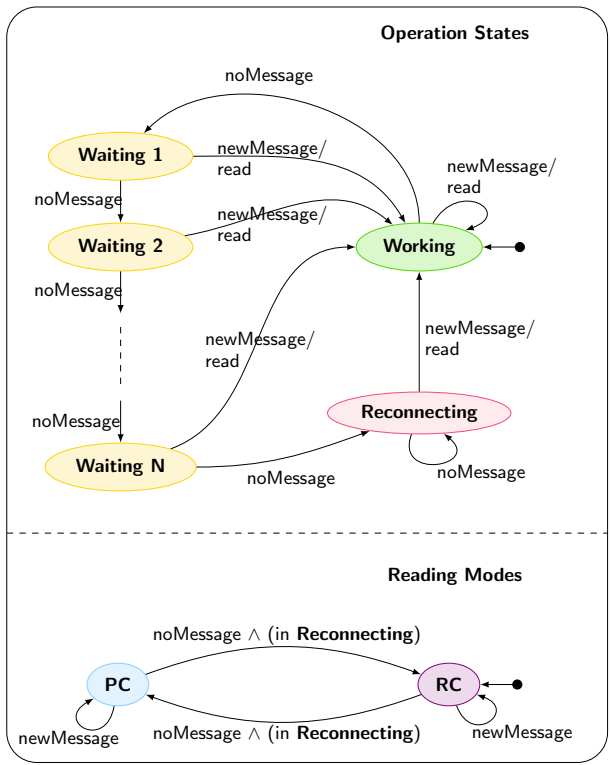

In particular, \cmpNoSep{Main Control} must implement two modes for reading messages: from the PC and from the RC.
If \cmpNoSep{Main Control} is in a certain reading mode and does not receive a message, set-points received in the previous control cycle will be processed. 
During $N$ cycles, \cmpNoSep{Main Control} will try to read a new message; if it fails, an alternation of reading modes, between the PC and the RC, will take place until a new message arrives.
This functionality is described by means of two concurrent Finite State Machines (FSM) in Figure \ref{fig:opStateFSM}.

The algorithms for reading messages from the PC and the RC have the same general structure. 
That is, both algorithms follow the same steps, but the implementations of some of these steps differ from each other---the PC connects via a serial protocol and the RC connects via a RF protocol.

\subsubsection{The usual design}

Figures \ref{fig:badIdea} illustrates the source code implementing the most usual design for the functionality described above. 
As can be seen, such a design establishes a single function with different nested \texttt{if} statements.
In the figure, \texttt{modePC} saves the reading mode and  \texttt{existentMessage} registers whether a new message  was read or not.
In turn, \texttt{countNoMsg} counts the number of failed readings; its value determines the alternation between reading modes when it reaches \texttt{N}. 
Thus, these variables record the system state that certain actions  depend on.
The described implementation could be part of  \mdNoSep{MainController} (Figure \ref{fig:diagGral}).

This approach results in repeated code and conditional sentences with, at least, three levels of nesting. 
How easily reusable and modifiable is this code? For example, what if one or more new reading modes, with different behaviours associated with them, have to be implemented? What if more conditions have to be considered in order to determine the corresponding behaviour? Under these circumstances, even more  conditional sentences should be added.  The question would then be, how many conditional sentences could be added, preserving legibility, changeability and testability? It has been shown that this kind of software structures results in a complex code, very difficult to modify and test \cite{Parnas1972}. Reuse is unfeasible as conditional branches cannot be called independently of each other.

\begin{figure}
\centering
\tikzumlset{font=\footnotesize\ttfamily}
\begin{tikzpicture}
\node(algo){};

\umlnote[width=.99\linewidth]{algo}{

1.\hspace{0.2cm}if mode==PC\\
2.\hspace{0.7cm}\textit{...to read message from PC...}\\
3.   \hspace{0.7cm}if existentMessage==True\\
4.      \hspace{1.2cm}countNoMsg=0\\
5.      \hspace{1.2cm}\textit{...to control with new set-points...}\\
6.   \hspace{0.7cm}else\\
7.      \hspace{1.2cm}countNoMsg = countNoMsg + 1\\
8.      \hspace{1.2cm}\textit{...to control with previous set-points...}\\
9.            \hspace{1.2cm}if countNoMsg >= N\\
10.               \hspace{1.7cm}mode=CR\\
11.\hspace{0.2cm}else\\
12.      \hspace{0.7cm}\textit{...to read message from RC...}\\
13.      \hspace{0.7cm}if existentMessage==True\\
14.            \hspace{1.2cm}countNoMsg=0\\
15.      \hspace{1.2cm}\textit{...to control with new set-points...}\\
16.      \hspace{0.7cm}else\\
17.      \hspace{1.2cm}countNoMsg = countNoMsg + 1\\
18.      \hspace{1cm}\textit{...to control with previous set-points...}\\
19.      \hspace{1.2cm}if countNoMsg >= N\\
20.         \hspace{1.7cm}mode=PC\\
21.\textit{...to write measurements and calculations...}
}
\end{tikzpicture}
\caption{Usual design but difficult to modify and reuse}
\label{fig:badIdea}
\end{figure}

\subsubsection{A design based on patterns}

An alternative in order to obtain a more flexible code is to apply Design Patterns. The proposed design is based on the State and the Template Method patterns \cite{Gamma:1995:DPE:186897}.  

The State pattern should be applied whenever a component's ``behaviour depends on its state, and it must change its behaviour at run-time depending
on that state'' \cite{Gamma:1995:DPE:186897}. 
The behaviour depicted in Figure \ref{fig:badIdea}, and attributed to \mdNoSep{MainController}, fits exactly in the purpose of State.
 Hence, the State pattern should be the first design choice for that functionality. This pattern will allow \mdNoSep{MainController} to change its behaviour when the reading mode changes (PC/RC) or when reading a new message is not possible. 

The State Pattern proposes to represent each state of Figure \ref{fig:opStateFSM} as a module, which will implement the behaviour associated with the corresponding state. \mdNoSep{MainController} will save internally a state, which may change under certain conditions; and specific actions will be delegated to the state. Besides, each state will know which is its successor state and will be responsible for carrying out the state change in \mdNoSep{MainController}. 

\begin{figure*}[h!]
{
\centering
\scalebox{0.8}{
\begin{tikzpicture}\hypertarget{fig:integral}{}

\umlclass{MainController}{}
{changeMode(Mode mode)\\
getMode() : Mode\\
changeOpState(OperationState st)\\
read()\\
}

\umlclass[type=abstract, above right=2cm and 2.5cm of MainController]{Mode}{}
{\umlvirt{newMessage():Bool}\\
\umlvirt{read()}\\
\umlvirt{changeMode}(MainController mCtrl)
}

\umlsimpleclass[below left=1.5cm and -1.5cm of Mode]{PC}

\umlclass[below right=1.5cm and -4cm of Mode]{RC}{}
{newMessage():Bool\\
read()\\
changeMode(MainController mCtrl)\\
}
\umlVHVinherit[arm2=-2]{PC}{Mode}
\umlVHVinherit[arm2=-2]{RC}{Mode}

\umluniaggreg[geometry=-|-, arm1=4.5cm, arg1=md, pos1=2.5]{MainController}{Mode}

\umlclass[type=abstract, below right=1.5cm and 3cm of MainController]{OperationState}{}
{read(MainController mCtrl)\\
\umlvirt{actionWithMsg(MainController mCtrl, Mode md)}\\
\umlvirt{actionNoMsg(MainController mCtrl)}\\
\umlvirt{control()}\\
\umlvirt{write()}
}
\umluniaggreg[geometry=-|-, arm1=4.5cm,arg1=opSt, pos1=2.5]{MainController}{OperationState}

\umlsimpleclass[below left =1.5cm and 6cm of OperationState]{Waiting}
\umlsimpleclass[below left =1.5cm and 2.5cm of OperationState]{WaitingN}

\umlsimpleclass[below left =1.5cm and -2cm of OperationState]{Reconnecting}
\umlclass[below right =1.5cm and -5cm of OperationState]{Working}{}
{actionWithMsg(MainController mCtrl, Mode md)\\
actionNoMsg(MainController mCtrl)\\
control()\\
write()
}

\umlVHVinherit[arm2=-2.8cm,anchor1=147]{Waiting}{OperationState}
\umlVHVinherit[arm2=-2.8cm, anchor1=144]{Reconnecting}{OperationState}
\umlVHVinherit[arm2=-2.8cm]{WaitingN}{OperationState}
\umlVHVinherit[arm2=-2.8cm]{Working}{OperationState}

\umlnote[above= 0.5cm of MainController, width=7cm]{MainController}{
\ttfamily 
md : Mode\\
opSt : OperationState\\
---------------------------\\
changeMode(Mode mode)\{\\
\hspace{0.5cm} md=mode\hspace{0.3cm}
\}\\
---------------------------\\
getMode()\{
\hspace{0.1cm} return md\hspace{0.3cm}
\}\\
---------------------------\\
changeOpState(OperationState st)\{\\
\hspace{0.5cm} opSt=st\hspace{0.3cm}
\}\\
---------------------------\\
read()\{
\hspace{0.1cm} opSt.read(this)\hspace{0.3cm}
\}\\
}

\umlnote[above right= 0.5cm and -1.5cm of RC, width=6.9cm, geometry=|-]{RC}{
\ttfamily 
modePC : PC\\
----------------------------\\
newMessage()\{\\
\textit{...is a new message available?...}
\}\\
----------------------------\\
read()\{\\
\textit{...to read message from the RC...}
\}\\
----------------------------\\
changeMode(MainController mCtrl)\{\\
\hspace{0.5cm} mCtrl.changeMode(modePC) \hspace{0.3cm}
\}
}

\umlnote[left=2.2cm of OperationState, width=6.1cm, anchor1=-17, anchor2=194]{OperationState}{
\ttfamily 
md1: Mode\\
------------------------\\
read(MainController mCtrl)\{\\
\hspace{0.5cm}md1 = mCtrl.getMode()\\\vspace{0.3cm}
\hspace{0.5cm}if md.newMessage()\\
\hspace{1cm}actionWithMsg(mCtrl,md1)\\
\hspace{0.5cm}else\\
\hspace{1cm}actionNoMsg(mCtrl)
\}
}

\umlnote[below left=-1.5cm and 1.3cm of Working, width=10cm, geometry=-|]{Working}{
\ttfamily 
wait1St : Waiting\\
--------------------------------\\
actionWithMsg(MainController mCtrl,Mode md)\{\\
\hspace{0.5cm}md.read() \hspace{0.3cm}\}\\
------------------------------------------\\
actionNoMsg(MainController mCtrl,Mode md)\{\\
\hspace{0.5cm}mCtrl.changeOpState(wait1St) \hspace{0.3cm}
\}\\
------------------------------------------\\
control(MainController mCtrl)\{\\
\textit{...control with new set-points...}\}
}

\end{tikzpicture}
}}
\caption{Design based on patterns}
\label{fig:integral}
\end{figure*}
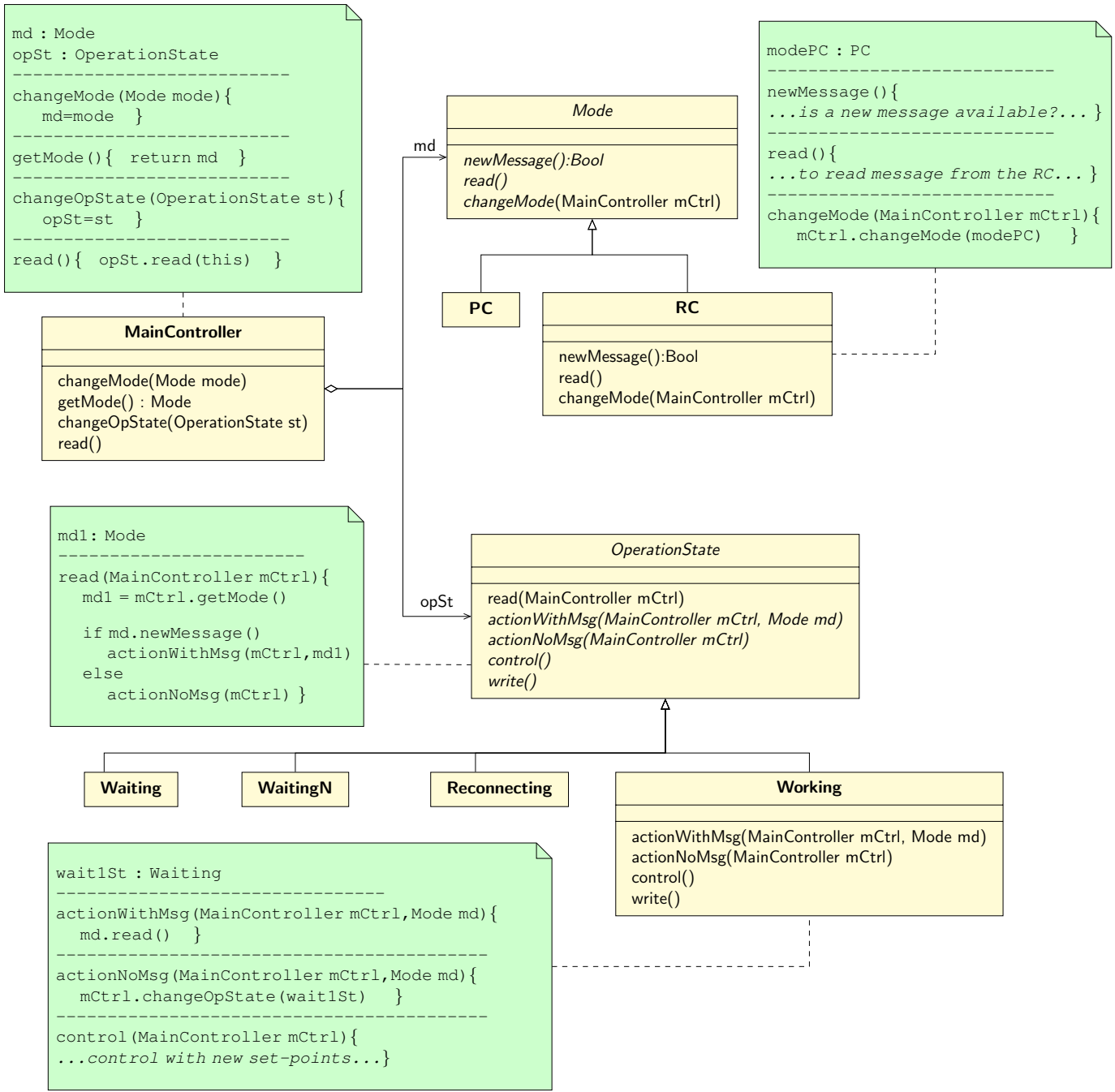

As there are two FSM describing the states of \mdNoSep{MainController}, there are two applications of State. 
Thus, as Figure \ref{fig:integral} illustrates, \mdNoSep{MainController} is composed of an instance of \abstNoSep{OperationState} (i.e., \texttt{opSt}) implementing an operation state (top of Figure \ref{fig:opStateFSM}) and an instance of \abstNoSep{Mode} (i.e., \texttt{md}) implementing a state corresponding to a reading mode (bottom of Figure \ref{fig:opStateFSM}). We will focus on the FSM describing the reading modes.

In Figure \ref{fig:integral}, \abstNoSep{Mode} is an interface\footnote{Names of interfaces and abstract methods are written in \textsf{\textit{sans-serif italics}}.} 
 common to the states relative to the reading modes. This interface declares methods in order to read a message. 
Module \mdNoSep{RC} inherits this interface and implements its methods in such way that a message coming from the RC can be read. Analogously, \mdNoSep{PC} implements its methods in order to read a message coming from the PC. 
Both modules provide the same methods, but their implementations are different. 

\mdNoSep{MainController} saves an instance of type \abstNoSep{Mode} (a \mdNoSep{PC} or  \mdNoSep{RC} instance), but will not know which one is in particular.
Consequently, in order to read a message, methods of \abstNoSep{Mode} will be invoked (\mdAbstMt{Mode}{newMessage} and \mdAbstMt{Mode}{read}). 
 Besides, \mdNoSep{MainController} will provide the method \mtNoSep{changeMode} in order to modify its reading mode.
Figure \ref{fig:integral} illustrates how this change is carried out by means of \mdAbstMt{Mode}{changeMode}, which invokes \mdMt{MainController}{changeMode} passing it as an argument the other reading mode (e.g., \texttt{mCtrl.changeMode(modePC)} in \mdMt{RC}{changeMode})\footnote{In the figures, the \texttt{pseudo-code} is written in \texttt{typewriter font}.}.

\noindent

If a new reading mode is necessary (e.g., from a Bluetooth device) only a new heir of \abstNoSep{Mode} implementing the corresponding functionality should be defined. The modules already implemented and  tested are not affected. This is an instance of the application of the open-closed design principle and is the result of applying the State pattern.

This design considerably simplifies building a product line. For example, if a robot model connects only with a PC, other model with a PC and a Bluetooth device and a third one connects with a PC and a RC, the images of all these three programs can be easily built by a convenient makefile. Each image will contain just the necessary modules thus generating smaller binaries. Conversely, the design described in Figure \ref{fig:badIdea} would require conditional compilation (which is not a best practice) or would not generate minimal images, thus possibly compromising hardware resources.

One more advantage of this approach is that if engineers managed to express the functionality as a FSM (e.g. Figure \ref{fig:opStateFSM}), then there is a standard, proven design based on State that engineers can rely on. In other words, State bridges the gap between the functional or behavioural view of \mdNoSep{MainController} and its module-based view.

The states and transitions concerning the FSM at the top of Figure \ref{fig:opStateFSM} are implemented in a similar fashion by applying the State pattern once more. 
Thus, \abstNoSep{OperationState} (Figure \ref{fig:integral}) defines an interface common to all operation states. Besides, since the main steps to read a message are similar in those states, the implementations of \mdAbstMt{OperationState}{read} are similar. That is, in any operation state, to read a message entails determining if there is a new message and then, to act accordingly (e.g., to read the message or to change the operation state). 
The Template Method pattern allows to define the skeleton of an algorithm, delegating certain steps of the algorithm to the submodules \cite{Gamma:1995:DPE:186897}. Thus, the latter can redefine some steps, whereas the algorithm structure is kept.
Thereby, Template Method should be the first design choice for message reading in the different operation states. 
In Figure \ref{fig:integral}, \mdAbstMt{OperationState}{read} is the \textit{template method} defining the structure of the algorithm to read messages (see \texttt{read}'s pseudo-code). This is the only one method implemented in \abstNoSep{OperationState}. Submodules inherit this implementation.
The method obtains the reading \abstNoSep{Mode}  (\texttt{md=mCtrl.getMode()}) and delegates the task of determining the availability of a message in \abstNoSep{Mode} (\texttt{md.newMessage()}). 
If there exists a new message, the template method calls \abstNoSep{OperationState::actionWithMsg}; otherwise, it calls \abstNoSep{OperationState::actionNoMsg}. Both methods are implemented in the submodules according to the required behaviour in each operation state.
Therefore, the actions carried out by  \mtNoSep{read} in \mdNoSep{Working}, \mdNoSep{Waiting}, \mdNoSep{WaitingN} and  \mdNoSep{Reconnecting} are different.

To sum up, in order to read a message, \mdMt{MainController}{read} delegates this task in the template method \mdAbstMt{OperationState}{read} (Figure \ref{fig:integral}). The latter, in turn, delegates the execution of \mtNoSep{actionWithMsg} and \mtNoSep{actionNoMsg} in the heirs of \abstNoSep{OperationState}. 
This delegation pattern introduces several points where new functionality can be added without affecting existing modules. It realises the open-closed design principle.

For example, it would be possible to add a new state called  \mdNoSep{WaitingK} which is reached after  $K-1$ attempts of reading a message. In this state the system could issue a warning message indicating that there is a potential connection  problem, although it will wait for $N$ attempts to move to the \mdNoSep{Reconnecting} state. The addition of this new state does not impact on the implementation of \mdAbstMt{OperationState}{read}.

After applying State twice and Template Method once, lines 1-3, 6, 11-13, 16 in Figure \ref{fig:badIdea} become the implementations of \mtNoSep{newMessage} and \mtNoSep{read} in the heirs of \abstNoSep{Mode}; and 
lines 4-5, 7-10, 14-15, 17-20 become the implementations of \mtNoSep{actionWithMsg} and \mtNoSep{actionNoMsg} in the heirs of \abstNoSep{OperationState}. 
Nested conditional statements are no longer necessary, while the number of lines of code is basically the same in both designs.

The design uses 8 design patterns, applied 30 times (Table \ref{tab:patternsResume} and Appendix \ref{appendix}), summing up 133 modules.

\begin{table}
\centering
\caption{Applied design patterns}
\label{tab:patternsResume}
\begin{tabularx}{\columnwidth}{Xc}
\toprule
\textsc{Number of design modules} & 133\\
\midrule
\textsc{Design Pattern} & \textsc{Applications} \\\midrule
Command & 7\\
State & 5\\
Strategy & 5\\
Decorator & 4\\
Builder & 4\\
Template Method & 3\\
Singleton & 1\\
Serializer & 1 \\\midrule
\textsc{Total} & 30 \\\bottomrule
\end{tabularx}
\end{table}

\section{Design Documentation}

The State pattern has been applied to design the MCU reading modes (PC/RC) (section \ref{sec:detailedDsgn}). However, neither diagrams as Figure \ref{fig:integral} nor their  implementation allow to easily identify   \mdNoSep{PC} or \mdNoSep{RC} as states of the State pattern. Thus, if reading messages from a Bluetooth device is required, although the design supports this requirement, developers may implement it violating conceptual design decisions. Eventually, this will turn the implementation into spaghetti code. 
This happens because reconstructing the fact that the implementation is based on State by just looking at the code, is difficult.
Additionally, after some time, design decisions, dependencies among modules or their functionalities can be forgotten. Rebuilding that knowledge from source code demands time and effort and is error-prone. Consequently, counting on good documentation is very convenient to understand and analyse software.

\begin{figure*}
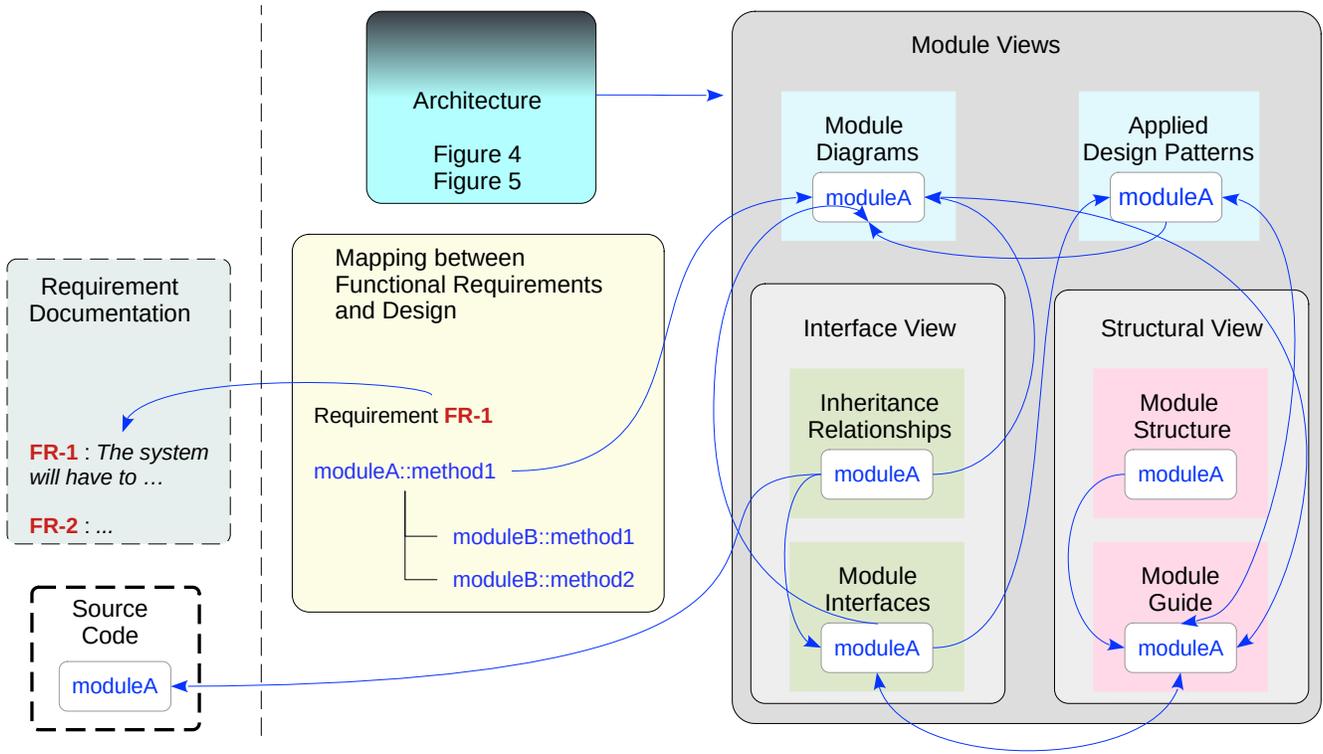

\centering
\pgfuseimage{documentation}
\caption{Design documentation at the right (of the dashed vertical line) and other documentation at the left. Blue arrows indicate hyperlinks.}
\label{fig:docSoftware}
\end{figure*}

Documentation of the MCU software design is based on \cite{ClementsEtAl2010}. Figure \ref{fig:docSoftware} depicts the documents generated in this project. As can be seen, different software views (architecture, detailed design, requirements, source code and so on) can be browsed through hyperlinks (blue arrows). The complete documentation is available in a public repository \cite{WRobotMCU}.

Next, through simple examples, a brief description of each view and its purpose is presented.

\subsection{Application of Design Pattern}

As explained above, the application of a design pattern must be documented in a way that  supported changes can be easily implemented. We use the \textbf{PatternApp} box to document the application of a given design pattern.

\noindent\begin{minipage}{0.95\columnwidth}
\begin{patternFTemp}{Reading modes}{ModeP}{integral}
\basedOn{State Pattern}
\why{Depending of the operation mode in which the system is, different reading mechanisms are realised.[...]}
\assigns
\is{\rmgTemp{Mode}{RC}}{State}
\is{\rmgTemp{PC}{RC}}{Concrete State}
\is{\rmgTemp{CR}{RC}}{Concrete State}
\is{\rmgTemp{Working}{RC}}{Context}
\is{\rmgTemp{WaitingN}{RC}}{Context}
\is{\rmgTemp{WaitingMAX}{RC}}{Context}
\is{\rmgTemp{Reconnecting}{RC}}{Context}
\end{patternFTemp}
\end{minipage}

\newcommand{\bFMaxi}{
\begin{tikzpicture}
\sf
\tikzstyle{botonF}=[shading=ball, ball color=blue!30!white, rectangle, rounded corners=2, font=\scriptsize]
\draw node[botonF]{F};
\end{tikzpicture}
}
\newcommand{\linkBFTempMaxi}[1]{\hyperlink{fig:#1}{\textcolor{black}{\bFMaxi}}}

In this case, the box documents the application of the State pattern to the reading modes. Why the pattern is applied and which could be likely changes are described. 
Besides, the role of each module, w.r.t. the pattern documentation, is specified by binding the module with an element of the pattern (`\textbf{where}' and `\textbf{is}' in the box). 
Thus, if a Bluetooth device is a new required reading mode, developers can clearly see that they can add a new State to implement the requirement. The \raisebox{-2pt}{\linkBFTempMaxi{ModeP}} button links the \textbf{PatternApp} box with a module diagram where the participating modules appear. In this way, users can navigate the documentation to understand the system from different perspectives.

\subsection{Interfaces and Inheritance}
The interface view documents the public methods of each module and, whenever necessary, the inheritance relation among them (Interface View in Figure \ref{fig:docSoftware}).
Thus, if an interface is modified, affected modules can be easily detected.  

The boxes below document the interfaces of \mdNoSep{Mode} and \mdNoSep{RC}, and the inheritance relation between them. The reserved words \textbf{Module}, \textbf{exports} and \textbf{comments} allow to describe name, interface methods and comments corresponding to a module. Besides, interface inheritance is also described through  `\textbf{inherits from}' (Interface Inheritances in Figure \ref{fig:docSoftware}). See how the navigation buttons work.

\noindent\begin{minipage}{0.95\columnwidth}
\begin{moduleFDPTemp}{Mode}{integral}{\linkBDPTemp{ModeP}{}}
\exports
\proc{\sf\textit{newMessage():Bool}}
\proc{\sf\textit{read()}}
\proc{\sf\textit{changeMode(MainController mCtrl)}}
\commTemp{Abstract module that provides an interface to receive orders coming from different sources.}
\end{moduleFDPTemp}
\end{minipage}

\noindent\begin{minipage}{0.95\columnwidth}
\begin{hmoduleFDPTemp}{RC}{Mode}{integral}{\linkBDPTemp{ModeP}{}}
\exports
\proc{\sf newMessage():Bool}
\proc{\sf read()}
\proc{\sf changeMode(MainController mCtrl)}
\commTemp{Concrete module that implements the methods to read orders coming from the RC.}
\end{hmoduleFDPTemp}
\end{minipage}

\subsection{Module Structure}

Modules should be organized hierarchically so developers can easily identify different portions of the system. This view is called \textit{structural} or \textit{decomposition} view.

In particular, we organise the modules in such way that a node in the  Module Structure, called \textit{logic module}, represents  
the minimal set of modules that a developer needs to know in order to implement a minimal functional unit. 
The aim of our decomposition is to assign portions of the implementation to independent development teams. 

As an example, the following module structure  describes \abstNoSep{Mode}, \mdNoSep{RC} and \mdNoSep{PC} as part of a logic module called \scNoSep{OperationModes}.

\begin{tikzpicture}[ grow via three points={one child at (0.5,-0.7) and
two children at (0.5,-0.7) and (0.5,-1.4)}, edge from parent path={(\tikzparentnode.south) |- (\tikzchildnode.west)}]
\tikzstyle{every node}=[anchor=west]
\tikzstyle{logic}=[font=\scshape\small]
\tikzstyle{abstract}=[font=\itshape\small]
\tikzstyle{concrete}=[font=\bf\small]
\node[logic](MCU){\sf\rmg{MCU}}

child{node[logic](OperationModes){\rmgTemp{\textsc{OperationModes}}{RC}}
		child{node[abstract]{\rmgTemp{\textit{Mode}}{RC}}}
		child{node[concrete]{\rmgTemp{RC}{RC}}}		
		child{node[concrete]{\rmgTemp{PC}{RC}}}
		}
child{node[below right =2.7cm and 0.1cm of MCU]{\small(... other modules...)}}
;
\end{tikzpicture}

The structural view should be augmented by documenting the responsibility, functionality and secret of each module. This view is called Module Guide. Developers use this information to implement and maintain the program. 
The following text box presents part of the documentation of \mdNoSep{RC} in the Module Guide.

\bigskip
\begin{tabular}{|p{0.9\columnwidth}|}
\hline
\begin{minipage}{0.85\columnwidth}
\begin{guidemFDPTemp}{\subsecTemp}{RC}{integral}{\linkBDPTemp{ModeP}{}}
\textit{Function}\\
$[$... description of the responsibilities of the module and its methods...$]$\\

[... specification of each method ...]\\
\textsf{read} receives a \mdNoSep{\rmgTemp{MainController}{RC}} and reads an order. [...]\vspace{0.2cm}\\
\hspace*{1cm}\begin{BVerbatim}
read(MainController mCtrl){
	mData=mCtrl.getData()
	mData.saveModeId(0)
	md.read(mCrtl)
}
\end{BVerbatim}

\textit{Secret}\\
$[$... description of the module secrets...$]$
\end{guidemFDPTemp}
\end{minipage}\\ 
\hline
\end{tabular}

\subsection{Mapping requirements to/from design elements}

Requirements traceability refers to how the design meets the requirements and how changes in the latter impact on the former \cite{ClementsEtAl2010}. 
Mapping each functional requirement to sequences of invocations among modules is useful to analyse how requirement changes affect the design and which are the involved modules. Besides, this mapping helps to verify that each requirement is satisfied.
For example, the  box below shows part of a requirement.

\smallskip
\noindent\fbox{
\begin{minipage}{0.95\columnwidth}\hypertarget{fr:ejReq}{}
\small
\textbf{FR-15.} \textit{The system shall set a waiting state if no message can be read.}
\end{minipage}
}
\smallskip

\textbf{FR-15} should be linked to the corresponding sequence of method invocations.
We use trees of method invocations in order to facilitate to go through different documents by means of hyperlinks, as follows.

\smallskip
\noindent\fbox{
\begin{minipage}{0.95\columnwidth}
\scriptsize
\hfill\\
\textbf{Requirement} \hyperlink{fr:ejReq}{\textcolor{red}{FR-15}}

\begin{useTreeS}
\sf
\mmRootTemp{MainController}{read}{integral}
child{\mmCBR{Working}{read}{integral}{-1.4}{-1.3}{MainController}
	child{\mmCBR{Working}{noMessage}{integral}{-0.9}{-0.2}{MainController}
			child{\mmCBR{MainController}{changeOpState}{integral}{-0.5}{1.2}{MainController}
		}
	}}
child{node[below right=0.3cm and -1.3cm of MainController]{...}}	
\end{useTreeS}
\end{minipage}
}\\

\section{Empirical validation}
Modifiability, reusability and maintainability are desirable quality attributes in software systems but they might compromise performance. 
Architectural styles and design patterns tend to generate chains of delegations between modules, which tends to increase the execution time.
One might even think that performance decreases to a point of making the software unusable. For this reason the performance of the implementation of our design needs to be empirically evaluated.

The MCU software has been implemented to run on a Kinetis K64 with an ARM Cortex-M4 core of 120 MHz and 256 KB of SRAM.  
The implementation was carried out using the MCUXpresso Suite of Software and Tools \cite{MCUXpresso}. 
The chosen programming language is C++.
The implementation results in around 12 KLOC distributed in 248 files of source code, available in a public repository \cite{WRobotMCU}.

\begin{table}[b]
\centering
\caption{Empirical evaluation of the performance of the MCU software}
\label{tab:performance}
\begin{tabularx}{\linewidth}{>{\raggedright}Xcr}
\toprule
\multicolumn{2}{l}{\textsc{Number of control cycles per scenario}} & 1000 \\
\multicolumn{2}{l}{\textsc{Available time in a control cycle}} & 100 \ms \\
\midrule
 & \multicolumn{2}{c}{\textsc{Time} (\ms)} \\
\multicolumn{1}{c}{\textsc{Description of execution scenarios}} & \textsc{Avg} & \textsc{Max} \\
\midrule
During the first 500 cycles, the wheel velocity setpoints are equal to 0 \rpm; after that, they are increased to 50 \rpm . & 2.12 & 3.28 \\\addlinespace[2pt]
During all cycles, the wheel velocity setpoints are equal to 50 \rpm while the steering setpoints constantly go back and forth between the extreme left and right positions. & 2.39 & 3.43 \\\addlinespace[2pt]
The wheel velocity setpoints are increased by 50 \rpm from 0 \rpm \ to 300 \rpm; after that, they are decreased by the same value until 0. & 2.65 & 3.59 \\\addlinespace[2pt]
The wheel velocity setpoints are as in the previous test while the steering setpoints constantly go back and forth between the extreme left and right positions. & 2.67 & 3.72 
\\\bottomrule
\end{tabularx}
\end{table}

Experts in Control Theory working with the WR determined that a 100 \ms control cycle is adequate to accomplish a suitable reaction time. Consequently, the MCU software has 100 \ms to receive orders coming from the PC or CR, read values from sensors, carry out control and, finally, send the resulting information to the PC. 

Hence, we defined four execution scenarios to analyse the performance of the software, as described in Table \ref{tab:performance}. Each scenario consists of processing a series of orders during 1000 control cycles.
The results show that the average time varied from 2.12 \ms to 2.67 \ms with a maximum of 3.72 \ms. It is noteworthy that these values are far from the 100 \ms available to perform the control tasks. Therefore, the empirical results permit to affirm that a design based on architectural styles and design patterns aiming at modifiability, reusability and maintainability can generate an implementation efficient enough as to accomplish its functions.

\section{Conclusions}

We have presented an approach to design robotic software, in which architectural styles and design patterns play a principal role. Particularly, we have shown the architecture and design of  the software of the MCU of a weeding robot. The design produces more flexible code that can easily incorporate likely changes. Besides, we have proposed a way of structuring the design documentation that allows to link requirements with design, and that makes it easier to  analyse the software in order to adapt it to likely changes.
Furthermore, we have shown that the implementation of this design features a performance that do not compromise the expected response time of the software.

In summary, our work shows that using architectural styles and design patterns in robotic software does not necessarily harm its performance and can provide reusability and modifiability.

\bibliographystyle{ieeetr} 
\bibliography{./referencias}

\newpage
\onecolumn
\appendix[Design Patterns Used in the MCU Software]
\label{appendix}

The following table lists all the design patterns used for the MCU software. Some of the design patterns are applied more than once to accommodate different requirements and anticipated changes. All design patterns are described by Gamma et al. \cite{Gamma:1995:DPE:186897}, except Serializer that is described by Martin et al. \cite{10.5555/273448}.

\begin{longtable}%
      {l>{\raggedright}p{4cm}>{\raggedright\arraybackslash}p{10cm}}
\toprule
\multicolumn{1}{c}{\textsc{Pattern}} & 
\multicolumn{1}{c}{\textsc{Pattern Application}} & 
\multicolumn{1}{c}{\textsc{Description and Anticipated Changes}}
\\\midrule\endhead
 & PC-MCU connection & 
The communication mode between the PC and the MCU depends on the states of connection between them. It could be necessary to add new connection states or to modify the behaviour associated with an existing state.
 \\\addlinespace[5pt]
 & Switch on and switch off the steering device & The actions necessary to enable or disable the steering device could change. 
 \\\addlinespace[5pt]
 & Operation states of the steering device controller&The steering device controller carries out different actions depending on the state of the system. Modifications, extensions or eliminations of the states or new actions on new states could be necessary.
 \\\addlinespace[5pt]
State & Reading modes&The mechanism by means of which an order is read, depends on whether the system is processing orders from the PC or from the RC. These mechanisms could change requiring new operation modes for receiving orders.
 \\\addlinespace[5pt]
 & Operation states of the main controller&Depending on its operation state (working, reconnecting, waiting, etc.), the main controller has a certain behaviour and indicates particular actions to the sub-controllers.
The behaviour associated with a state could change or new operation states could be required. 
\\\midrule
 & Control algorithms for wheels & A wheel movement command can be given by a tension, current or velocity order. Depending on this, different control algorithms calculate the tension to apply to a wheel. These algorithms could change or a new algorithm could be necessary. 
 \\\addlinespace[5pt]
 & Control algorithm for the steering device & The control algorithm that determines the turn direction of the steering device could change. 
 \\\addlinespace[5pt]
Strategy & Functions & The system needs to carry out different calculations (e.g, the RC signals are calculated from some inputs). These calculations could change or new calculations could be necessary. 
 \\\addlinespace[5pt]
 & Orders for wheel control system & The main controller orders the wheel control sub-systems to execute commands (e.g., to register the wheel position or to read a value measured by a sensor). The way in which each order is carried out could change or new orders could be required. 
 \\\addlinespace[5pt]
 & Algorithms for wheel behaviour &The system requires different algorithms in order to carry out wheel controlled movements (e.g., advance, stop, reverse). These algorithms could undergo modifications or new algorithms could be needed. \\\midrule
Singleton & Constant values & Constant values that are required by more than one module, are hidden in a \textit{singleton} in order to guarantee the uniqueness of values.\\\midrule
 & Function arguments& The calculation functions require a same subset of basic arguments, but they differ in other ones.
Thus, arguments can be added or eliminated depending on what a function needs.
 \\\addlinespace[5pt]
Decorator & Collectors of time instants & Time instants must be registered to calculate values relative to signals coming from the RC and the Hall sensors.
The registered time instants are used to carry out different calculations which are extensions of a basic calculation. New calculations or calculation modifications could be necessary. 
 \\\addlinespace[5pt]
 & Algorithm extensions to establish the wheels' behaviour & Some algorithms that allow to carry out wheel movements are the result of combining or extending simpler algorithms (e.g., reversing). Changes or new extensions of these algorithms could be needed. 
 \\\addlinespace[5pt]
 & Operation modes as extensions of a basic task & The operation mode of the system is determined by the source of the orders; that is, the PC or the CR. In both operation modes, reading an order involves the same subset of basic actions which are extended with other actions. How an operation mode reads orders could change, as well as a new reading mode could be required. \\\midrule
 & RC Buffers & The RC signals are received through a velocity pin and a direction pin. Each pin is associated with a buffer which registers and interprets signals by using an algorithm. Each algorithm performs the same steps but they differ in how they carry out each step (e.g. signal interpretations). Changes in the signal interpretations could arise, due to a new RC model, or the interpretation of a new signal could be necessary.
 \\\addlinespace[5pt]
Template method & Main controller orders for control sub-systems & The main controller gives the same sequence of commands to all control sub-systems (e.g., to read sensed values, to control, etc.). However, each subsystem carries out each command in a particular way. These distinctive features could change or a new order could be necessary.
 \\\addlinespace[5pt]
 & Read orders in the operation states of the main controller & The operation state (e.g. working, waiting, etc.) of the main controller depends on the possibility of reading a new order from the PC or the CR. In all states, reading an order entails the same sequence of steps, but how a certain step is carried out is determined by each particular state. The way in which a certain step is realised could change.
\\\midrule
 & Commands to control wheel behaviour & Stopping a wheel, setting its tension or changing its orientation are orders that can have effects on different system elements. The actions to realise the orders or the involved elements could change.
 \\\addlinespace[5pt]
 & Commands to control steering device behaviour & Disabling the steering device, determining its turn direction, etc. entails actions carried out on different elements of the system.  The elements or such actions could change.
 \\\addlinespace[5pt]
 & Process Hall sensor interruptions (callback substitution) & An interruption coming from a Hall sensor involves actions on certain elements of the system. The actions or  the elements could change. 
 \\\addlinespace[5pt]
Command & Process timer interruptions for the steering system (callback substitution) & From a timer interruption,  actions on elements of the steering device are carried out. These actions could require modifications or the elements could change.
 \\\addlinespace[5pt]
 & Process timer interruptions for the wheel control system (callback substitution) & Actions are executed on certain components of the wheel control system, when a timer interruption takes place. Such actions or components could change.
 \\\addlinespace[5pt]
 & Process the RC pin interruptions (callback substitution) &
 A interruption coming from a RC pin entails actions on some system components. The actions could require modifications or the components affected by the actions could change.
 \\\addlinespace[5pt]
 & Process timer interruptions for the main controller (callback substitution) & Interruptions coming from a timer mark the beginning of each control cycle. Thus, when an interruption arises, the system must carry out certain actions. These actions or the affected modules could change.
 \\\midrule
Serializer & Receiving and sending information between the MCU and the PC or the RC & The MCU interchanges data with the PC or the RC. The way of accessing to the data and its format could change and, a new data source or destination could be necessary. 
 \\\midrule
 & Building a wheel control system & A wheel control system is composed of a set of parts that relate to each other in a particular way.
 These parts and the relations among them could require modifications.  \\\addlinespace[5pt]
Builder & Building a steering device control system & Building a steering device control system entails to combine different software pieces. The group of pieces involved and the mode in which they related to each other could change. 
 \\\addlinespace[5pt]
 & Building a control system pool & The control system pool managed by the main controller could require to incorporate new control systems .
 \\\addlinespace[5pt]
 & Building a main controller & Main controller is composed of different elements. The set of these elements and their relations could be modified (e.g., if a new device is added, the main controller must incorporate new elements in order to control it). 
 \\\bottomrule
\end{longtable}

\end{document}